\documentclass[letterpaper]{article} 
\usepackage{aaai25}  
\usepackage{times}  
\usepackage{helvet}  
\usepackage{courier}  
\usepackage[hyphens]{url}  
\usepackage{graphicx} 
\usepackage{natbib}  
\usepackage{caption} 
\usepackage{algorithm}
\usepackage{amsmath}
\usepackage{algorithmic}
\usepackage{subcaption}
\usepackage{xcolor}
\usepackage{booktabs}
\usepackage{multirow}
\usepackage{multicol}
\usepackage{paralist}

\newcommand{\blu}[1]{\textcolor{black}{#1}}
\usepackage{newfloat}
\usepackage{listings}
\DeclareCaptionStyle{ruled}{labelfont=normalfont,labelsep=colon,strut=off} 
\floatstyle{ruled}
\newfloat{listing}{tb}{lst}{}
\floatname{listing}{Listing}
\title{Exploring Disparity-Accuracy Trade-offs in Face Recognition Systems:\\The Role of Datasets, Architectures, and Loss Functions}

\author {
    Siddharth D Jaiswal\textsuperscript{\rm 1},
    Sagnik Basu\textsuperscript{\rm 1},
    Sandipan Sikdar\textsuperscript{\rm 2},
    Animesh Mukherjee\textsuperscript{\rm 1}
}
\affiliations{
    \textsuperscript{\rm 1}Indian Institute of Technology, Kharagpur, India\\
    \textsuperscript{\rm 2}L3S Research Centre, Leibniz University, Hannover, Germany
}

\usepackage{bibentry}

\begin{document}

\maketitle

\begin{abstract}
Automated Face Recognition Systems (FRSs), developed using deep learning models, are deployed worldwide for identity verification and facial attribute analysis. The performance of these models is determined by a complex interdependence among the model architecture, optimization/loss function and datasets. Although FRSs have surpassed human-level accuracy, they continue to be disparate against certain demographics. Due to the ubiquity of applications, it is extremely important to understand the impact of the three components-- model architecture, loss function and face image dataset on the accuracy-disparity trade-off to design better, unbiased platforms. In this work, we perform an in-depth analysis of three FRSs for the task of gender prediction, with various architectural modifications resulting in ten deep-learning models coupled with four loss functions and benchmark them on \blu{seven face datasets across 266}
evaluation configurations. Our results show that all three components have an individual as well as a combined impact on both accuracy and disparity. We identify that datasets have an inherent property that causes them to perform similarly across models, independent of the choice of loss functions. Moreover, the choice of dataset determines the model's perceived bias-- the same model reports bias in opposite directions for \blu{three} gender-balanced datasets of ``in-the-wild'' face images of popular individuals. Studying the facial embeddings shows that the models are unable to generalize a uniform definition of what constitutes a ``female face'' as opposed to a ``male face'', due to dataset diversity. We provide recommendations to model developers on using our study as a blueprint for model development and subsequent deployment\footnote{\textcolor{red}{This work has been accepted for publication at ICWSM 2025.}}.
\end{abstract}

\section{Introduction}
\label{sec:intro}

\begin{figure}[!t]
    \centering
    \includegraphics[width=0.55\linewidth, keepaspectratio]{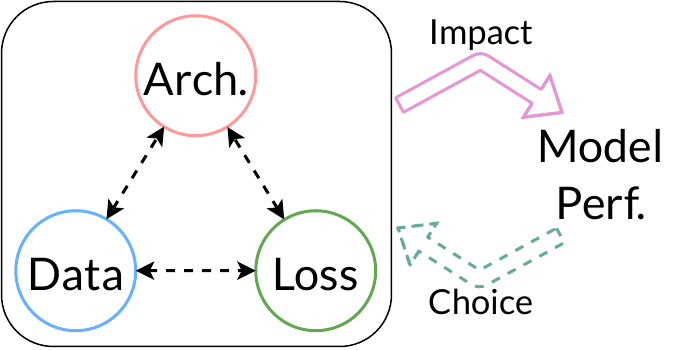}
    \caption{Interdependence between the architecture, data and loss function that determine the performance of a model, which in turn informs the choice of components.}
    \label{fig:dependency}
\end{figure}

The performance of deep learning models is usually determined by a complex interdependency among the model architecture, the objective function being optimized for, and the data, as shown in Figure~\ref{fig:dependency}. For example, even a very deep architecture coupled with the optimal loss function cannot perform well if the data is not representative of the true population. Similarly, a model may not perform well on a perfectly sampled dataset if the architecture is too shallow or the loss function is too simple.
This is especially true for face recognition models, which have become highly commonplace in society with the development and democratization of deep learning and rival human-level accuracy for a number of tasks~\cite{o2021face}. These systems ingest very diverse data in the form of face images and attempt to find a general pattern to classify attributes like the perceived gender of the person~\cite{karkkainen2021fairface}. Use cases for such attribute classification range from targeted advertising~\cite{mennecke2013avatars} and customer analytics~\cite{reutersjapan2010} to surveillance~\cite{hitoshi2016video} and content moderation~\cite{ning2022face}. Despite the high accuracy of these classification models, disparities have been reported on commercial~\cite{buolamwini2018gender,jaiswal2022two} and open-source FRSs~\cite{jaiswal2024breaking} against minority demographics, severely impacting access to services\blu{~\cite{dunnepassport2019}}, leading to exclusion and unfair treatment. Such disparities are an outcome of the way the dataset, the model architecture and the objective function together behave.  

An FRS model can have multiple types of vision backbones viz. CNNs and transformers, use multiple types of loss functions~\cite{wang2018cosface,deng2019arcface} and be trained and tested on various datasets~\cite{raji2020saving,karkkainen2021fairface}. These components can be combined in an exponential number of ways, with each having an impact on the final accuracy and, more importantly, the disparity in performance amongst classes, especially for minority demographics. In this study, we attempt to unravel this complex relationship and its impact on gender bias in FRSs for the task of gender prediction through an in-depth, large-scale audit study involving three FRSs. The FRSs have two types of vision backbones and four types of loss functions in various combinations and are evaluated over \blu{seven} benchmark face datasets with more than \blu{86k} images in total. 

\noindent \textbf{Research questions:} 
It is usually seen that accuracy and bias are often competing objectives~\cite{janssen2021bias}. Thus, before deploying any FRS for sensitive tasks like gender prediction, it is absolutely imperative to evaluate how the three components -- data, model architecture, objective optimised -- impact the accuracy vs disparity divide. This brings us to our first research question-- \textbf{RQ1} \textit{Do the individual components, viz. model architecture, loss function and dataset, impact the disparity along with the accuracy?} 

Most studies in the literature show that a majority of FRSs are biased against females. In this work, we attempt to investigate whether this observation is generalizable across datasets, architectures and loss functions or if there exists a more nuanced relationship that determines the direction of disparity. We formalize this into our second research question-- \textbf{RQ2} \textit{Do all datasets report equally disparate performance against females, irrespective of the model architecture and loss function?} Through this question, we will be able to understand not only the extent of disparity but also the group that is more often discriminated against.

Disparities in FRSs exist because the model performs better for one sub-group over the others. Thus, independent of the dataset, it is important to quantify how the accuracies for each gender group change for different choices of loss functions and architectures. This brings us to our final research question -- \textbf{RQ3} \textit{What is the relationship between the change in accuracy for males vs. females for different models and their architectures?} Through this question, we seek to understand which models and loss functions prefer which gender group.

\noindent \textbf{Our key observations.} 
In line with the research questions highlighted above, we make the following key observations from our in-depth analysis.
\begin{compactitem}
    \item Our analysis of the accuracy vs disparity trade-off shows that model architecture, loss function and dataset impact not only the accuracy but also the observed disparity in an FRS, irrespective of the choice of backbone-- CNN or transformer. Our evaluation shows that independent of other model parameters, in the accuracy vs disparity chart, some datasets like CFD, \blu{LFW} and CelebSET always cluster around high accuracy with low disparity, while UTKFace and Fairface report low disparity but at the expense of low accuracy \blu{and CelebA reports high disparity \& low accuracy} (answers RQ1). 
    \item We observe that not all FRSs are equally disparate against females in all datasets. In fact, the CNN model LibfaceID reports lower accuracies for males on \blu{two standard balanced datasets-- CelebSET \& CelebA}. 
    With increasing model complexity, not only does the overall disparity reduce, but so does the effect of the loss function on the disparity, too (answers RQ2).
    \item We note that the two gender groups have different levels of sensitivity toward each model and architecture. For example, the accuracy simultaneously improves for females and degrades for males when residual connections are used in the LibfaceID model. Similarly, using just the CLS embedding in the vision transformer model elicits both positive and negative changes for the males' accuracy and primarily a positive change in accuracy for the females. (answers RQ3).  
\end{compactitem}
\section{\blu{Related Work}}
\label{sec:related}
We briefly discuss the existing literature for FRSs, their biases against minority or marginalized groups and existing studies focusing on the interplay among datasets, model architectures and loss functions.

\noindent\textbf{Face recognition systems.} These platforms are primarily based on deep learning models and target various tasks like face identification~\cite{yang2002detecting} and downstream analysis~\cite{levi2015age} like gender, age, emotion and face matching~\cite{taigman2014deepface}. 
The models can be commercial~\cite{aws_rekognition,facepp,microsoft_face} or open-source~\cite{taigman2014deepface,deng2019arcface,wang2018cosface}.
SOTA transformer models~\cite{chen2016supervised,dan2023transface,zhong2021face} are also used for similar tasks, with human-level accuracy.
Reduction in deployment costs and optimized models has allowed wide-scale adoption of these platforms at city-scale levels~\cite{livemint2021}.
Gender prediction on FRSs comes across as an unobvious task, but FRSs are rampantly used for this task in domains like targeted advertising~\cite{mennecke2013avatars}, customer analytics~\cite{reutersjapan2010} and  surveillance~\cite{hitoshi2016video}. 
This raises an important concern about the biases that can manifest while using FRSs as gender predictors. Since this is becoming highly normative, such systems need to be monitored, and the untoward biases need to be addressed.

\noindent\textbf{Biases in FRSs.} Multiple studies in prior literature have exposed social biases based on sensitive features like gender/race, etc., for the task of gender prediction in traditional CNN-based commercial~\cite{buolamwini2018gender,jaiswal2022two,raji2020saving} and open-source~\cite{jaiswal2024breaking} FRSs. While researchers have studied biases in vision transformers~\cite{liu2022contextual,brinkmann2023multidimensional}, these have been for a general set of tasks rather than the specific task of gender prediction from face images. As these vision transformers are highly complex models, and used for a wide variety of face recognition tasks, it is important to study and mitigate the biases therein.

\noindent\textbf{Interaction among the three main components of deep learning.} Existing literature has studied the interplay between the structure of data and loss functions~\cite{d2021interplay}, but these have not focused on the impact towards the overall bias in the system. \citet{davidian2024exploring} study the impact of dataset size and imbalance in CNNs for healthcare while~\citet{cherepanova2023deepdive} perform a similar study for Face Recognition Systems.~\citet{cabannes2023ssl} study the interplay amongst the choices for data augmentation, network architecture and training algorithm, but do not attempt to study the impact on the bias of a model. In this work, we not only study the impact of the choice of datasets, model architecture, and loss functions on the accuracy but also on the disparity in the model, a far more societally impactful problem.

\noindent\blu{\textbf{Relevance to web \& social media.} 
In this work, we study the fairness and bias implications of FRSs, which are deployed not only in the physical world but also on digital platforms. For example, FRSs have been deployed at a large scale on both Google Photos~\cite{googlephotosFRS} and Facebook (shut down in 2021~\cite{facebookphotosFRS1} and revived in 2024~\cite{facebookphotosFRS2}) for tagging and scammer detection. These services have also shown highly discriminatory biases~\cite{zhang2015google,bbc2021fb,bbc2021twitter} indicating the need to audit the models, the datasets and their frameworks in the context of social media applications. On the other hand, the social media research community has shown growing interest in this domain-- both in the use of these FRSs for analysing social media platforms~\cite{chakrabortyICWSM2017,VikatosHT2017,MessiasWI2017,PangBigData2015} as well as 
studying fairness concerns in image tagging~\cite{KyriakouICWSM2019,BarlasICWSM2019} and attribute analysis~\cite{jaiswal2022two,JungICWSM2018} applications. We argue that FRSs are highly prevalent in both physical and social networks, and it is important to study them end-to-end to ensure their deployment is indiscriminate and within regulatory frameworks.
With FRSs increasingly being deployed in web and social media applications, we believe our study to be both highly relevant and timely.
} 
\color{black}
\section{\blu{Datasets \& FRS Models}}
\label{sec:datasets-models}
In this section, we present a brief overview of the datasets and open-source FRS models that we audit in this study.

\subsection{Datasets}
\label{sec:datasets}

\begin{table}[!t]
    \centering
    \footnotesize
    \begin{tabular}{l r r r c}
    \toprule
    \multirow{2}{*}{\textbf{Name}} & \multicolumn{3}{c}{\textbf{\# Images}} & \multirow{2}{*}{\textbf{Usage}}\\
    \cmidrule{2-4}
    & \textbf{Male} & \textbf{Female} & \textbf{Total}\\
    \midrule
    Adience & 8,107 & 9,356 & 17,463 & Train/FT/Test\\
    \hline
    CFD & 680 & 761 & 1,441 & Test\\
    CelebSET & 800 & 800 & 1,600  & Test\\
    FARFace & 931 & 931 & 1,862 & Test\\
    \blu{LFW} & \blu{2,966} & \blu{2,966} & \blu{5,932} & \blu{Test}\\
    Fairface & 5,788 & 5,160 & 10,948 & Test\\
    UTKFace & 12,390 & 11,314 & 23,703 & Test\\
    \blu{CelebA} & \blu{20,500} & \blu{20,500} & \blu{41,000} & \blu{Test}\\
    \bottomrule
    \end{tabular}
    \caption{Summary characteristics of the benchmark datasets.}
    \label{tab:data} 
\end{table}

We consider a range of datasets of different sizes with significant variety in gender, race and geographic distributions. We use the Adience~\cite{eidinger2014age} dataset, with more than 17k images, to train the CNN model and to fine-tune the pre-trained transformer models. We use five diverse benchmark datasets for evaluation--
\begin{compactitem} 
    \item Chicago Face Database (CFD)~\cite{ma2015chicago}, curated from images volunteered by citizens of the USA, belonging to four different races-- White, Black, Asian and Latinx.  CFD has two extension sets-- CFD-MR~\cite{ma2020chicago}, with 88 images and CFD-India ~\cite{lakshmi2021india} with 146 images. We combine them with the original dataset and refer to the superset as CFD throughout.
    \item CelebSET~\cite{raji2020saving}, curated from IMDB images of Hollywood celebrities belonging to the White and Black races.
    \item FARFace~\cite{jaiswal2024breaking}, curated from ESPNCricInfo images of cricketers belonging to the Global North and Global South ($>50\%$). The authors do not annotate with the race labels.
    \item \blu{LFW~\cite{LFWTech}, curated from the web with names and gender of each individual. The authors do not annotate the faces with the race labels.} 
    \item Fairface~\cite{karkkainen2021fairface}, curated from the YFCC-100M Flickr dataset belonging to the following races -- White, Black, Latinx, Indian, Southeast Asian, East Asian and Middle Eastern.
    \item UTKFace~\cite{zhang2017age}, curated from the Morph dataset and the CACD dataset. This dataset has the following races -- White, Black, Asian, Indian and Others (including Hispanic, Latinx, and Middle Eastern). 
    \item \blu{CelebA~\cite{liu2015faceattributes}, curated from celebrity face images around the world collected from the internet. The authors provide 
    fine-grained annotation for gender and other facial attributes, but not race.}
\end{compactitem} 

All the datasets mentioned above are annotated with the binary gender -- male \& female. The labels in CFD were self-annotated by the individuals whose photos are part of the dataset, whereas for all other datasets, the labels were annotated by the dataset curators.
Consequently, our gender prediction models are also trained to predict a binary gender label. However, we do acknowledge that the notion of gender is fluid, and all predictions should be interpreted as the \textit{perceived} gender. \blu{A more detailed discussion is presented in the section on limitations. All datasets except CFD are curated from online sources and have played an important role in the training and deployment of FRSs in the physical world as well as social media platforms~\cite{taigman2014deepface}, thus further reiterating their close association with social media platforms. For example, Google uses FRSs to perform person re-identification for image tagging~\cite{googlephotosFRS}, and Facebook is using FRSs to identify scammers using celebrity images on their platform~\cite{facebookphotosFRS2}.}  
The dataset sizes, along with the distribution between the two binary gender labels, are noted in Table~\ref{tab:data}.

\subsection{\blu{FRS Models}}
\label{sec:models}
We audit two types of vision backbones -- CNNs and vision transformers, by evaluating three types of face recognition models -- LibfaceID~\cite{levi2015age}, ViT-Face~\cite{zhong2021face} and an instruction-tuned vision language model, InstructBLIP~\cite{dai2305instructblip}. The models differ in terms of architectural complexity, size of training data and parameter size, i.e., 8M to 7B. We present a brief description of the models as follows.

\begin{figure}
	\centering
	\begin{subfigure}{0.48\textwidth}
	\centering
		\includegraphics[height=2.5cm, keepaspectratio]{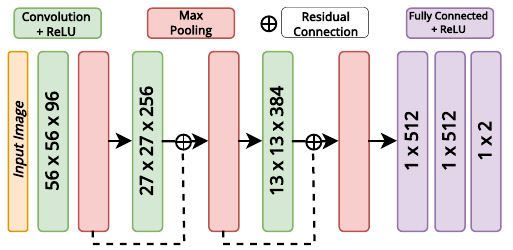}
		\caption{Residual connections with the base architecture}
		\label{fig:libfaceid-skip}
	\end{subfigure}\\
	\begin{subfigure}{0.48\textwidth}
	\centering
		\includegraphics[height=2.5cm, keepaspectratio]{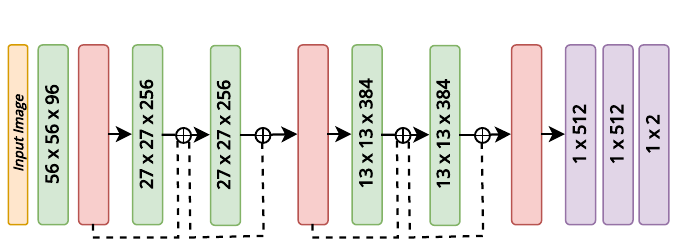}
		\caption{Residual connections with the deeper architecture}
		\label{fig:libfaceid-skip2}
	\end{subfigure}%
	\caption{Schematics for the Libfaceid~\cite{levi2015age} model modified only with residual connections (\ref{fig:libfaceid-skip}), and with extra layers and residual connections (\ref{fig:libfaceid-skip2}).} 
	\label{fig:libfaceid_models}
\end{figure}

\noindent $\bullet$~\textbf{{Libfaceid\footnote{\url{https://github.com/richmondu/libfaceid}}}~\cite{levi2015age}} -- We use the implementation based on a CNN model proposed by~\citet{levi2015age}.
\textcolor{black}{The model has a simple CNN backbone with three \texttt{Conv} and three \texttt{FC} layers. We train this model from scratch on the Adience~\cite{eidinger2014age} dataset. The training hyperparameters are in Table~\ref{tab:hyperparameters}, and other implementation details are present in the Appendix.}

\noindent\noindent $\bullet$~\textbf{{ViT-Face}~\cite{zhong2021face}} -- This model is based on the classic vision transformer~\cite{dosovitskiy2020image} backbone. The model is deeper and more complex than the CNN architecture and learns image representations by dividing it into patches and linearizing them. The model has an input patch size of 12 with a stride size of 8. The base model is simply trained to generate image embeddings. Hence, we add two linear layers to the final embedding layer to perform gender prediction. The model is fine-tuned on the Adience dataset. We refer to the model as ViT throughout.

\noindent $\bullet$~\textbf{{InstructBLIP}~\cite{dai2305instructblip}} -- 
InstructBLIP is an extended version of BLIP-2~\cite{li2023blip} vision-language model. It uses a frozen image encoder (ViT-g/14~\cite{fang2023eva}) and a frozen large language model (LLM) (Vicuna-7B~\cite{vicuna2023}) along with a QFormer~\cite{li2023blip} which connects the two frozen models and is pre-trained using task-specific instructions. Similar to the vision transformer above, we add two additional linear layers to the QFormer's output and fine-tune with the Adience dataset.

\section{\blu{Experimental Design}}
\label{sec:experiment}
We now describe our experimental setup for all the FRS models, using different loss functions along with the architectural changes to evaluate the accuracy and disparity for the task of gender prediction. 

\subsection{\blu{LibfaceID Settings}}
We first describe the architectural changes, followed by the loss functions that we use to train this model.

\subsubsection{Architectural changes}
The base LibfaceID CNN model is shallow with three convolution layers. We refer to this model as LBFC$_B$. Next, we extend this architecture with three changes, resulting in three new models-- (a)~residual connections between the \texttt{Conv} layers (see Fig.~\ref{fig:libfaceid-skip}) to carry forward information from previous layers (referred to as LBFC$_{B+R}$), (b)~two extra \texttt{Conv} layers to increase the model depth (referred to as LBFC$_{B+2}$) and, (c)~residual connections in the deeper model (see Fig.~\ref{fig:libfaceid-skip2}) (referred to as LBFC$_{B+2+R}$). 

For both models in Figure~\ref{fig:libfaceid_models}, we also experiment with an additional setup where the residual connections are weighted. Details and results for this setup are in the Appendix.

\subsubsection{Loss functions}
The default loss that we train the model with is the Cross-Entropy loss, referred to as ``CE'' in all our experiments. This is a commonly used classification loss function, deployed for its simplicity, especially for binary classification tasks. In this work, we also train the CNN model with two additional SOTA loss functions, explicitly introduced for face recognition models, with the aim of improving inter-class separability-- (a)~\textit{Triplet} loss~\cite{schroff2015facenet} (T), a contrastive loss function which requires a positive and negative example for every input (known as the anchor). We set the positive example as another image of the same gender and the negative example as an image from the opposite gender. (b)~\textit{ArcFace} loss~\cite{deng2019arcface} (A), an angular margin loss function that attempts to increase inter-class separability and intra-class compactness. 
These loss functions are used along with the CE loss for all experimental setups in isolation, as well as in combination-- ``T'', ``A'' and ``A+T''. The weight for each loss function is decided using grid search, with details in the Appendix.

\subsection{\blu{ViT-Face Settings}}
We do not modify the core architecture of ViT-Face; instead, we choose between different embeddings for fine-tuning the base model. 

\subsubsection{Choice of embeddings}
The vision transformer generates embeddings for each of the input patches and an extra embedding for the CLS token at the end of the pipeline.
We choose the following embeddings to fine-tune our model-- (a)~the embedding generated from the CLS token, referred to as ViT$_{CLS}$. (b)~the embedding generated by taking a mean of all patch embeddings-- ViT$_M$ and, (c)~the embedding generated by concatenating the previous two-- ViT$_{CLS+M}$.

\subsubsection{Loss functions}
Similar to the CNN model, here as well, we use CE as our default fine-tuning loss function. 
Next, we use the T and A loss functions described above. We also experiment with the SOTA \textit{CosFace}~\cite{wang2018cosface} (Cos) loss function, designed to use a cosine margin based function to increase the inter-class distance between different faces. In our experiments, we note that the combination of these losses works well only with ViT$_{M}$ and ViT$_{CLS+M}$. Thus, we use the loss functions individually for ViT$_{CLS}$ and in combination for the other two models-- ``CE'', ``T'', ``A'', ``Cos'', ``A+T'', ``A+T+Cos''. It must be noted that the CE loss is always a part of the pipeline with other loss functions. The weights for each combination are shared in the Appendix.

\subsection{\blu{InstructBLIP Settings}}
Similar to the vision transformer, we do not modify the core architecture, instead fine-tune it with two linear layers. As the vision-language model works best with textual prompts, we provide the following string as input with each query image-- \textit{``What is the gender of this person?''}. The choice of loss functions is the same as in ViT$_{M}$ and ViT$_{CLS+M}$. Further details on the weights for each combination are present in the Appendix.
\color{black}

\subsection{\blu{Evaluation Metrics}} 
The metric we use to evaluate the models is accuracy-- $Acc$ and the fairness metric is disparity between the accuracies for each gender group $Acc_M - Acc_F$, referred to as gender disparity. \blu{More formally, we define the two metrics as follows-- $Acc$ = \(\frac{C_M + C_F}{T_M + T_F}\) where $C_G$: $G \in \{M~(\text{male}), F~(\text{female})\}$ is the number of correct predictions for each gender group, and $T_G$ is the number of total data points for each gender group. Similarly, the disparity is formally defined as $Acc_M - Acc_F$ = \(\frac{C_M}{T_M}\) - \(\frac{C_F}{T_F}\)}. 

Overall, we consider \textit{ten} models, \textit{seven}
datasets and \textit{four} loss functions in various combinations, leading to a total of \textbf{\blu{266}}
evaluation configurations. \blu{The hyperparameters used in our experiments are noted in the Appendix.}
\begin{figure*}[!t]
	\centering
	\begin{subfigure}{0.25\textwidth}
        \centering
		\includegraphics[width= \textwidth, height=2.5cm, keepaspectratio]{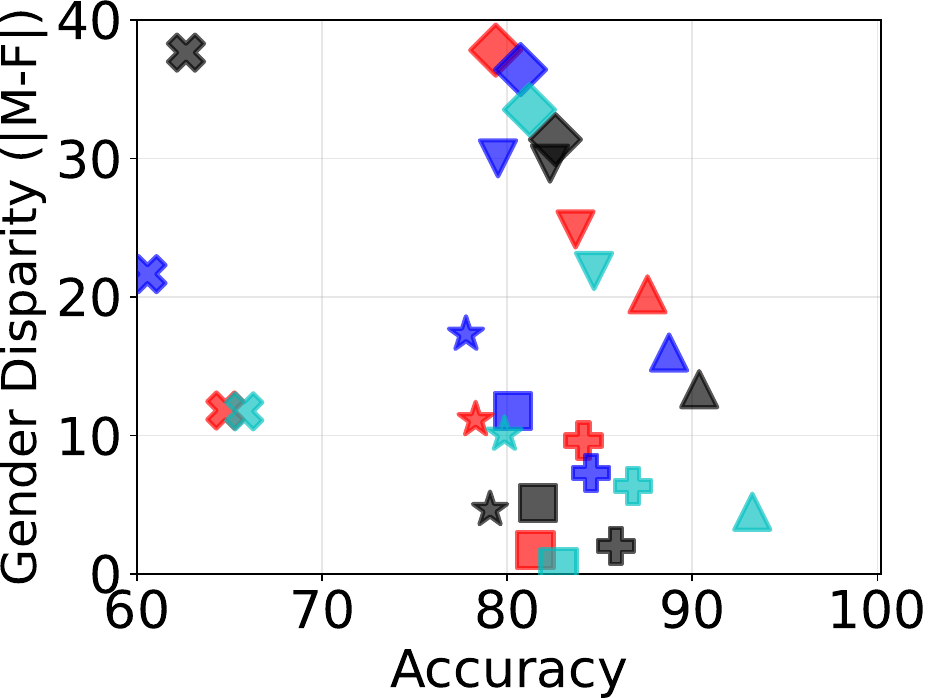}
        \caption{LBFC$_{B}$}
	\end{subfigure}%
	\begin{subfigure}{0.25\textwidth}
        \centering
		\includegraphics[width= \textwidth, height=2.5cm, keepaspectratio]{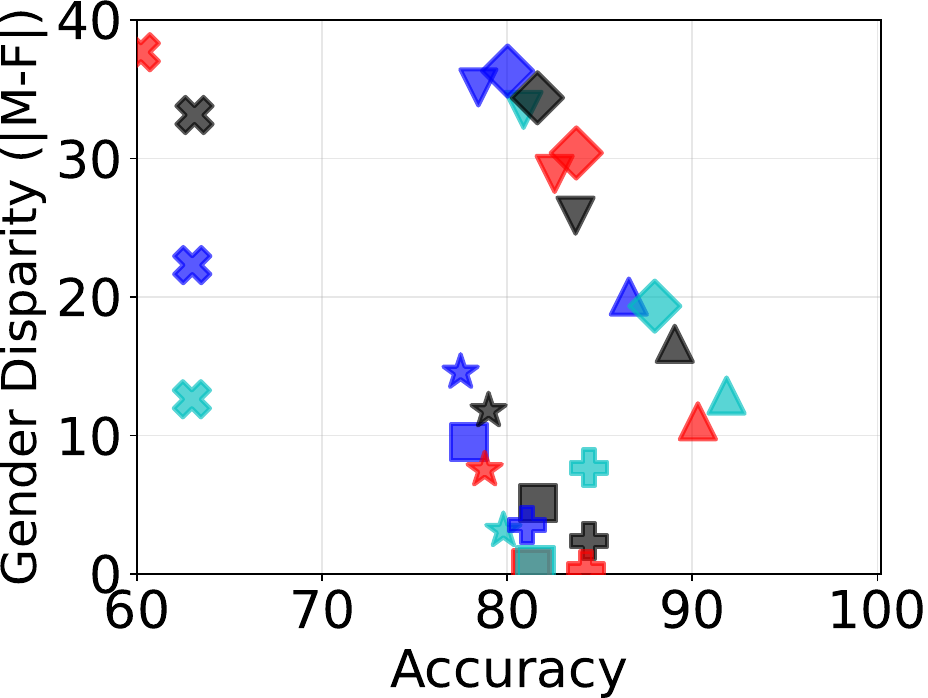}
        \caption{LBFC$_{B+R}$}
	\end{subfigure}%
	\begin{subfigure}{0.25\textwidth}
        \centering
		\includegraphics[width= \textwidth, height=2.5cm, keepaspectratio]{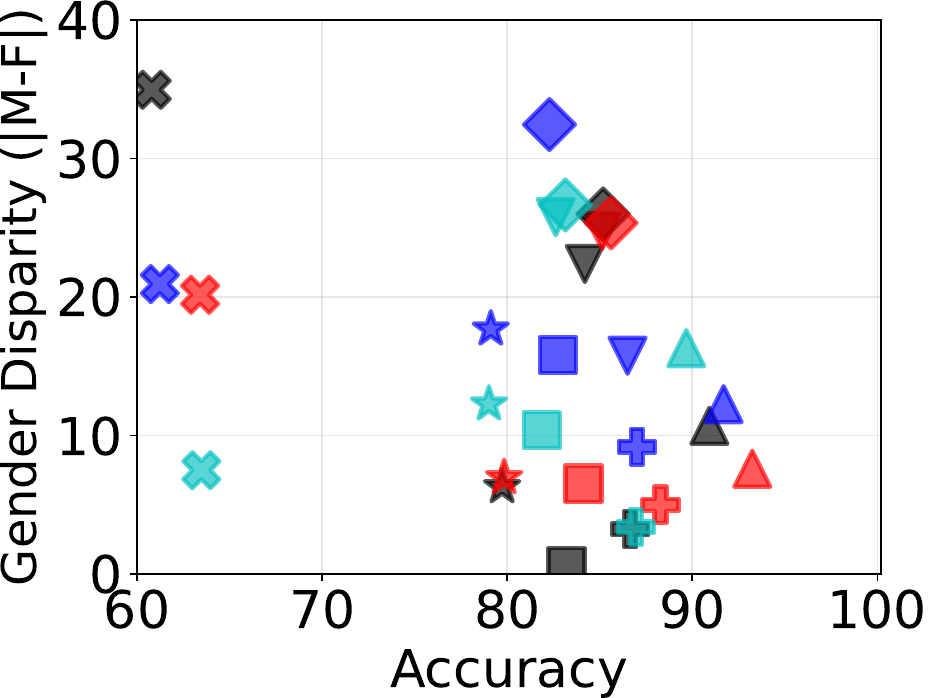}
        \caption{LBFC$_{B+2}$}
	\end{subfigure}%
	\begin{subfigure}{0.25\textwidth}
        \centering
		\includegraphics[width= \textwidth, height=2.5cm, keepaspectratio]{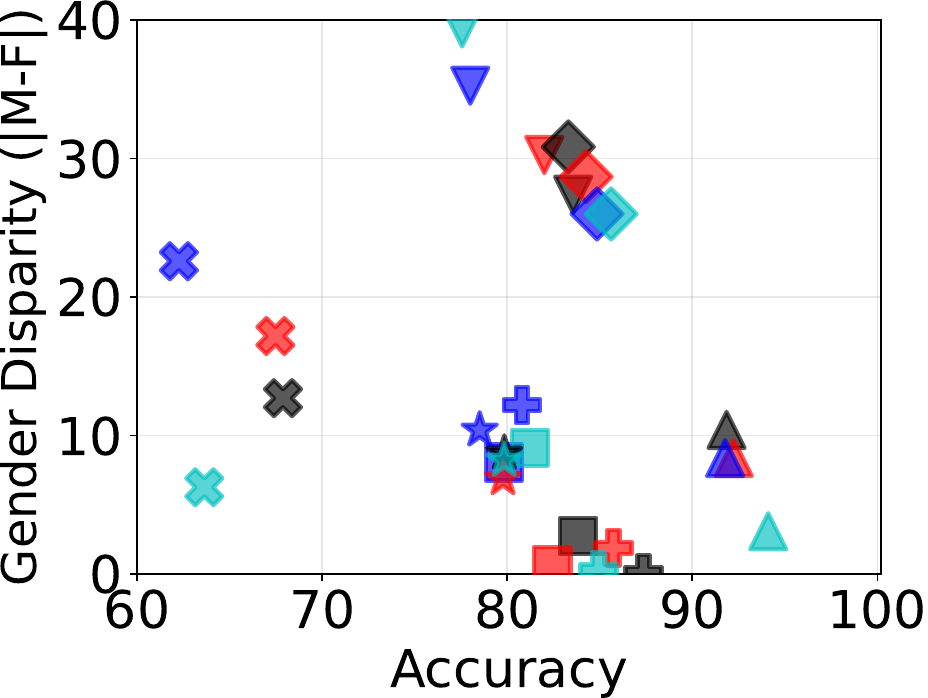}
        \caption{LBFC$_{B+2+R}$}
	\end{subfigure}%
\\
	\begin{subfigure}{0.25\textwidth}
        \centering
		\includegraphics[width= \textwidth, height=2.5cm, keepaspectratio]{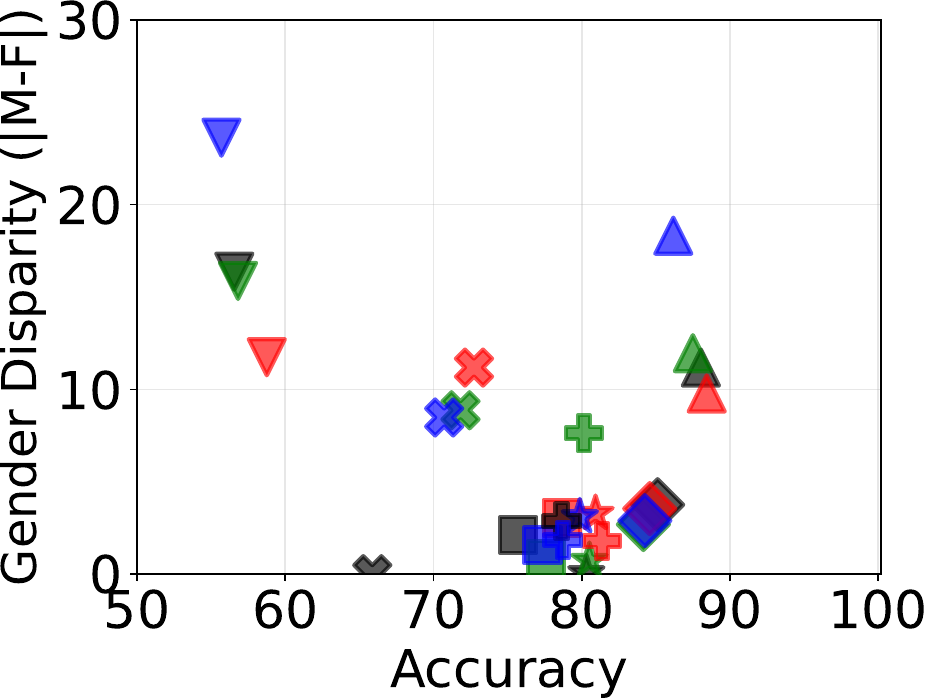}
        \caption{ViT$_{CLS}$}
	\end{subfigure}%
	\begin{subfigure}{0.25\textwidth}
        \centering
		\includegraphics[width= \textwidth, height=2.5cm, keepaspectratio]{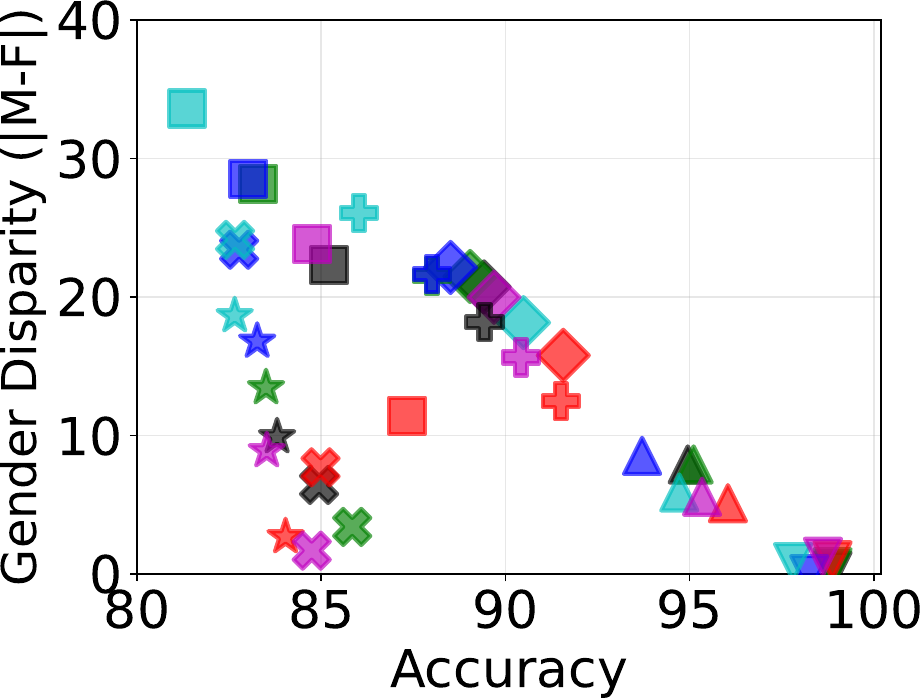}
        \caption{ViT$_{M}$}
	\end{subfigure}%
	\begin{subfigure}{0.25\textwidth}
        \centering
		\includegraphics[width= \textwidth, height=2.5cm, keepaspectratio]{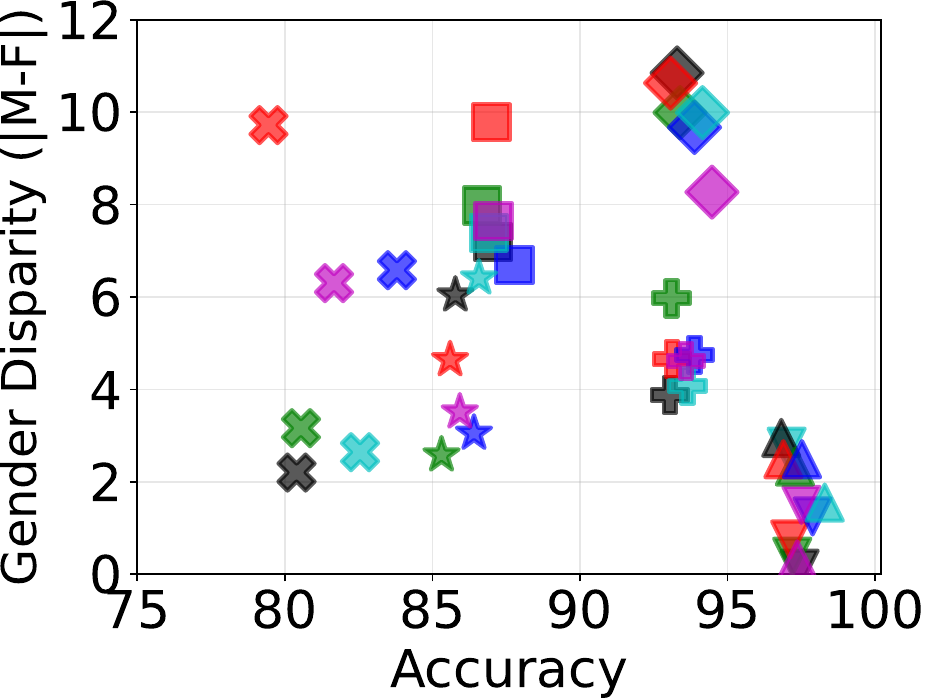}
        \caption{ViT$_{CLS+M}$}
	\end{subfigure}%
	\begin{subfigure}{0.25\textwidth}
        \centering
		\includegraphics[width= \textwidth, height=2.5cm, keepaspectratio]{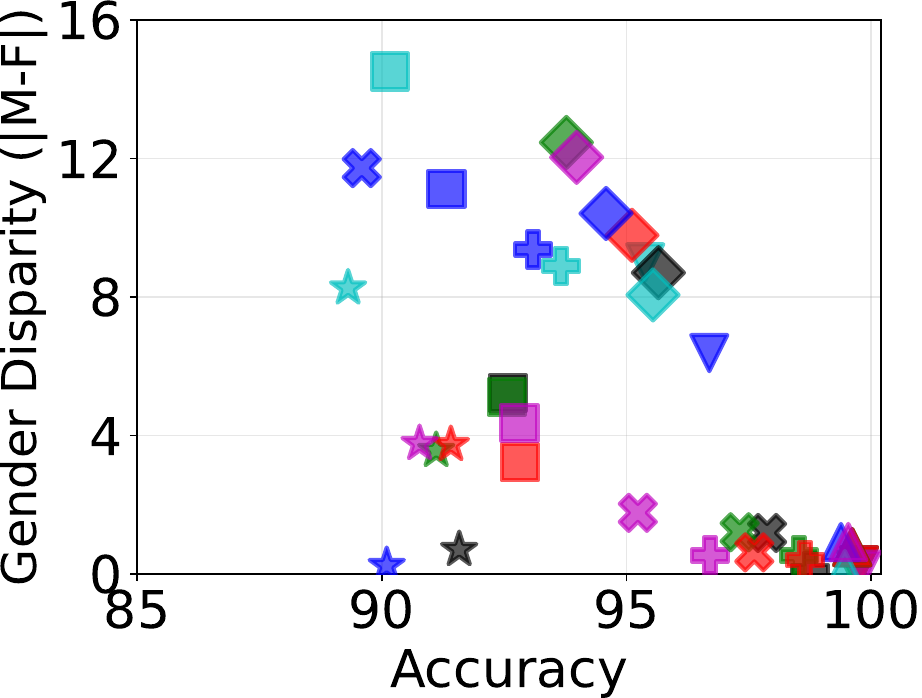}
        \caption{IBLIP}
	\end{subfigure}
    \\
    \begin{subfigure}{0.52\textwidth}
        \centering
        \includegraphics[width= \textwidth, height=2.5cm, keepaspectratio]{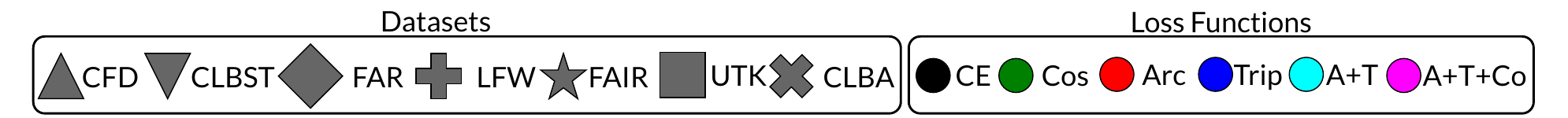}
    \end{subfigure}
	\caption{Accuracy vs. absolute disparity for different architectures across all datasets and loss functions. CFD always reports high accuracy, whereas FARFace \blu{and CelebA primarily} report high disparities. On every architecture, each dataset has a similar performance for all loss functions that it is evaluated with. The shapes refer to the datasets, with colors referring to the various loss functions. Each combination of dataset and loss function refers to one experimental setup. The model complexity, choice of loss function and choice of dataset impact both the accuracy and the disparity.
    }
	\label{fig:scatter-acc-disp}
\end{figure*}

\section{Results \& Observations}
\label{sec:results}
We now present the results and associated takeaways from our in-depth audit study. 
Here we only discuss the results of the test benchmark datasets; the results on the test subset for Adience (which was used for training/fine-tuning the models) are present in the Appendix. 
We reiterate here that the CNN model has been trained from scratch, whereas the transformer and VLM have been fine-tuned.

\subsection{\blu{Accuracy vs. Disparity (RQ1)}}

Accuracy and disparity are often competing objectives in a model's training procedure. Here, we attempt to get a detailed understanding of how the architecture, loss function and dataset combine to predict the gender of a person.  In Figure~\ref{fig:scatter-acc-disp}, we present the scatter plots between the accuracy and absolute disparity for each dataset, segregated by the model architecture type.

\subsubsection{Results for LibfaceID}
The LibfaceID model is a small CNN trained on the Adience dataset for the task of gender prediction. 
\blu{(a)~In Figs.~\ref{fig:scatter-acc-disp}a--d, we see that the accuracy is always between 60-95\% for all architectures, datasets and loss functions, and the highest absolute disparity is 40\% and below.} 
\blu{(b)~Next, we see that, across all architectures, Triplet loss and ArcFace loss compete for the highest disparity.} 
\blu{(c)~Some of the datasets always report low absolute disparities, whereas others always report high accuracy -- FARFace and CelebA report the highest disparity (except in Fig.~\ref{fig:scatter-acc-disp}d), whereas UTKFace has the lowest disparity. Similarly, CFD has the highest accuracy, and CelebA has the lowest accuracy.}
(d)~Finally, we see a strong clustering tendency amongst the datasets, especially when residual connections are used, with three primary forms of clusters -- (i) high accuracy, low disparity, (ii) low accuracy, low disparity and (iii) low accuracy, high disparity. \blu{The last type, reported primarily for the CelebA dataset, is the most adversarial input for the model and exposes its shortcomings.}

\subsubsection{Results for ViT}
The complex attention-based architecture of the pre-trained vision transformer is expected to improve the accuracy, but does not provide a guarantee on the reduction of disparity. 
(a)~We see that the accuracy is in a wide range -- between 55--90\% when only CLS embeddings are used as opposed to 80--98\% when the mean of all image patch embeddings is also used. 
\blu{The range of absolute disparity is largest in $ViT_{M}$ and smallest in $ViT_{CLS+M}$.}
(b)~Here, we note that the architecture plays a stronger role than the loss functions in determining the accuracy and disparity, with no clear trend across different loss functions. (c)~We also note that neither CFD reports the highest accuracy nor UTKFace consistently reports the lowest disparity, thus showing how sensitive each dataset is to the architecture. (d)~Finally, we see that the clustering tendency is even higher for this model, especially for $ViT_{CLS+M}$.

\subsubsection{Results for InstructBLIP}
The VLM includes a frozen language model and a Q-Former block that drives the vision encoder to focus on the salient parts of the image, thereby improving the accuracy of the task even further.
(a)~It is clear that this is the best-performing model with a high avg. accuracy -- $\geq 90\%$ and low avg. disparity -- $\leq 12\%$. (b)~Similar to LibfaceID, using only triplet loss reduces the accuracy for each dataset. \blu{(c)~FARFace reports the largest disparity (considering all loss functions), and Fairface reports the lowest accuracy. CFD and CelebSET are the best-performing datasets.} (d)~The clustering tendency is the highest, with a clear separation into the three types of clusters observed earlier.

\subsubsection{Major takeaways} 
We now list the major takeaways from the results in Figure~\ref{fig:scatter-acc-disp}.

\noindent $\bullet$ \textit{Model architectural complexity impacts performance--} As expected, LibfaceID is the worst-performing model, and InstructBLIP is the best-performing model in terms of both high accuracy and low disparity.

\noindent $\bullet$ \textit{Choice of loss function impacts performance--} 
Triplet loss has been used previously by researchers~\cite{jaiswal2024breaking} to improve the accuracy of FRSs for gender classification, but our experimental results show that auditing this loss function for different architectures and datasets exposes the shortcomings of this optimization.
Thus, for a fixed dataset and architecture, the choice of loss function has a significant impact on the performance (Fig.~\ref{fig:dependency}). 

\noindent $\bullet$ \textit{Choice of dataset impacts performance--} \blu{FARFace and CelebA most often report the largest disparity, whereas CFD reports the highest accuracy.} This is strongly correlated to the inherent nature of the datasets-- CFD is a highly standardized benchmark dataset with every person having the same pose, angle, lighting, and clothing, and FARFace is a new, recently released dataset that has a majority of the faces from the Global South, thus providing a new, adversarial challenge, even to large pre-trained models.

For a given architecture, the performances of a dataset for the different loss functions are very close to each other, considering both accuracy and disparity, thus creating a cluster. For example, see Fig.~\ref{fig:scatter-acc-disp}g, where each dataset's symbols for all losses are close to each other. A similar observation is true for the other architectures as well. \blu{The inherent property of the test dataset seems to strongly influence the fairness of the inference results.} 

Thus, we see that the three components of a model indeed have a significant impact on the model's performance, impacting both accuracy and disparity, answering RQ1. 

\noindent\blu{\textbf{Debiased algorithms}: To complete the analysis, we investigate the effect of the loss functions discussed earlier on known debiased FRSs. We choose the popular model provided by \citet{karkkainen2021fairface} for this purpose. The authors pretrained a ResNet-34 network with the Fairface dataset to ensure that it is debiased. We call this the vanilla version of the model. We then fine-tune the pretrained model with all the loss functions on the Adience dataset. From Table~\ref{tab:debiasing}, we see that the choice of loss functions positively impact both accuracy and disparity, with our choice of fine-tuning setup \textit{reducing the disparity on 5 out of the 6 datasets} (we ignore Fairface as the model is trained on the same set and has risk of data overlap).}

\begin{figure*}[!t]
	\centering
	\begin{subfigure}{0.25\textwidth}
        \centering
		\includegraphics[width= \textwidth, height=3.2cm, keepaspectratio]{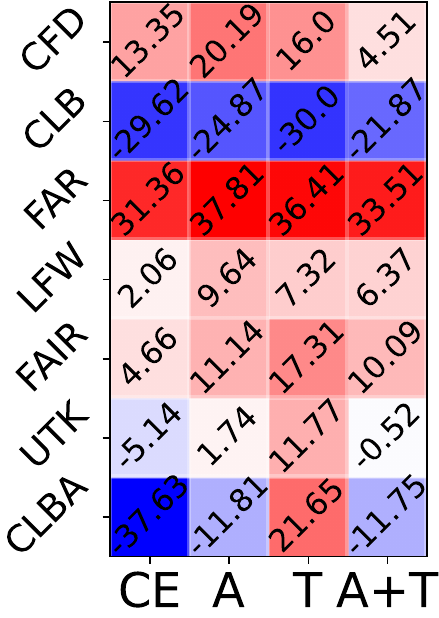}
        \caption{LBFC$_{B}$}
	\end{subfigure}%
	\begin{subfigure}{0.25\textwidth}
        \centering
		\includegraphics[width= \textwidth, height=3.2cm, keepaspectratio]{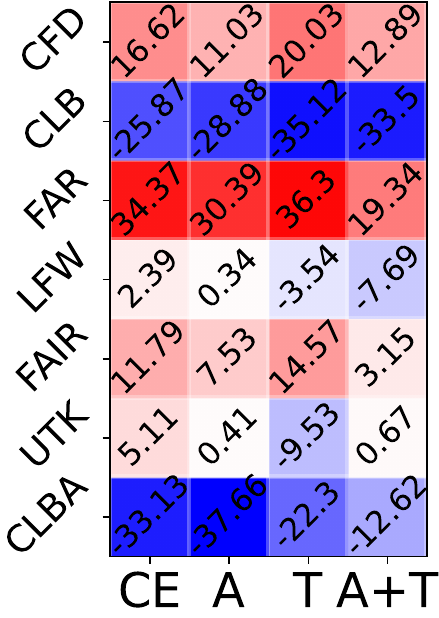}
        \caption{LBFC$_{B+R}$}
	\end{subfigure}%
	\begin{subfigure}{0.25\textwidth}
        \centering
		\includegraphics[width= \textwidth, height=3.2cm, keepaspectratio]{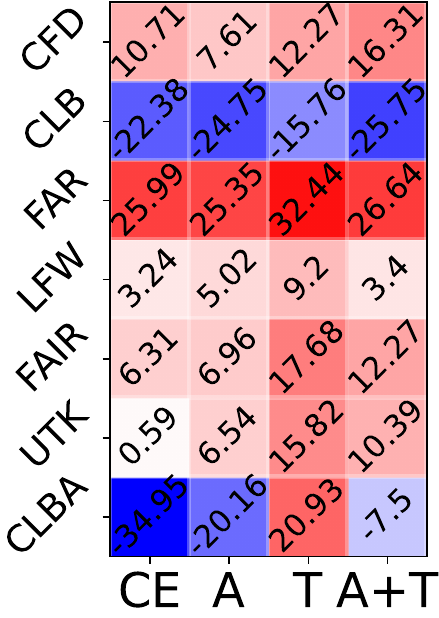}
        \caption{LBFC$_{B+2}$}
	\end{subfigure}%
	\begin{subfigure}{0.25\textwidth}
        \centering
		\includegraphics[width= \textwidth, height=3.2cm, keepaspectratio]{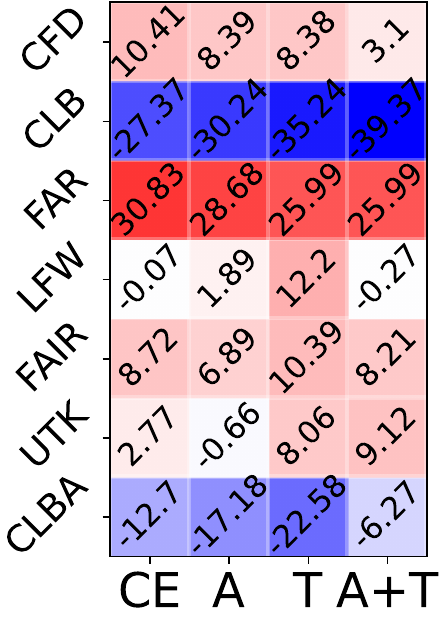}
        \caption{LBFC$_{B+2+R}$}
	\end{subfigure}%
\\
	\begin{subfigure}{0.25\textwidth}
        \centering
		\includegraphics[width= \textwidth, height=3.2cm, keepaspectratio]{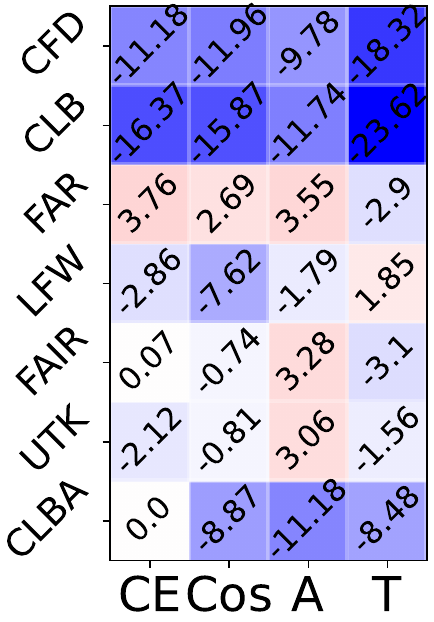}
        \caption{ViT$_{CLS}$}
	\end{subfigure}%
	\begin{subfigure}{0.25\textwidth}
        \centering
		\includegraphics[width= \textwidth, height=3.2cm, keepaspectratio]{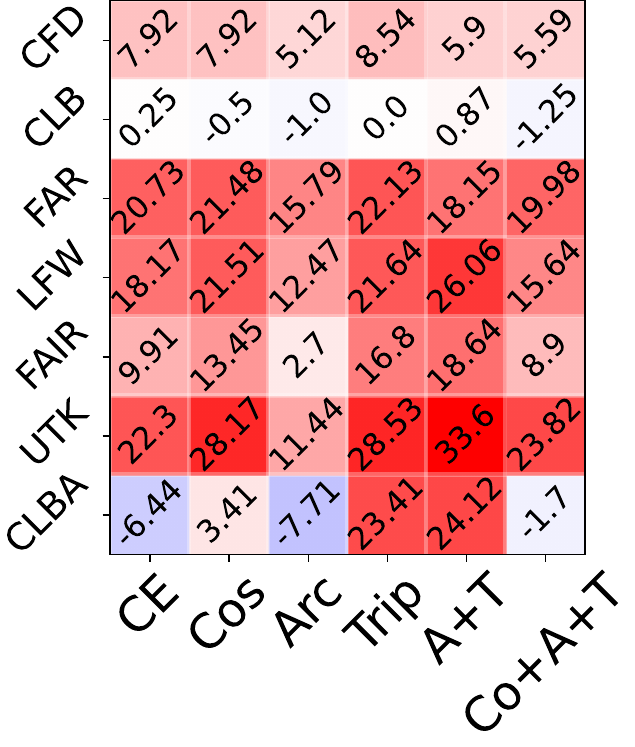}
        \caption{ViT$_{M}$}
	\end{subfigure}%
	\begin{subfigure}{0.25\textwidth}
        \centering
		\includegraphics[width= \textwidth, height=3.2cm, keepaspectratio]{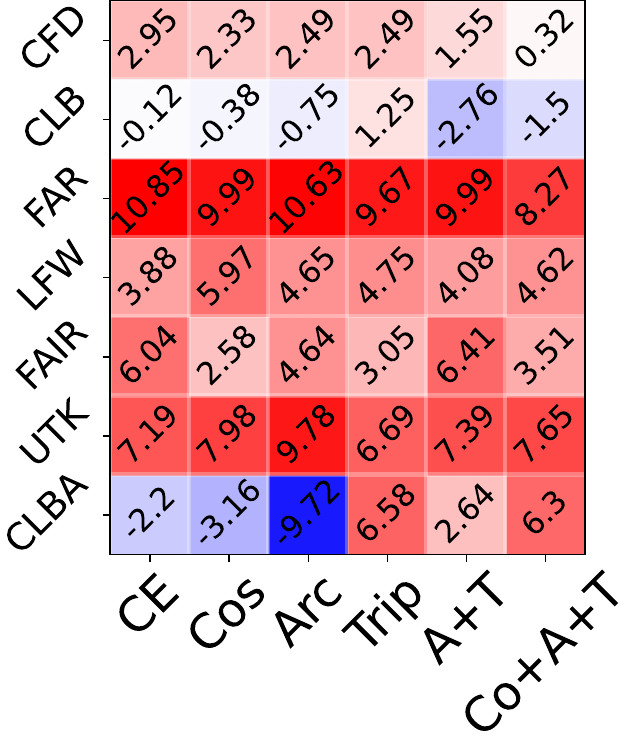}
        \caption{ViT$_{CLS+M}$}
	\end{subfigure}%
	\begin{subfigure}{0.25\textwidth}
        \centering
		\includegraphics[width= \textwidth, height=3.2cm, keepaspectratio]{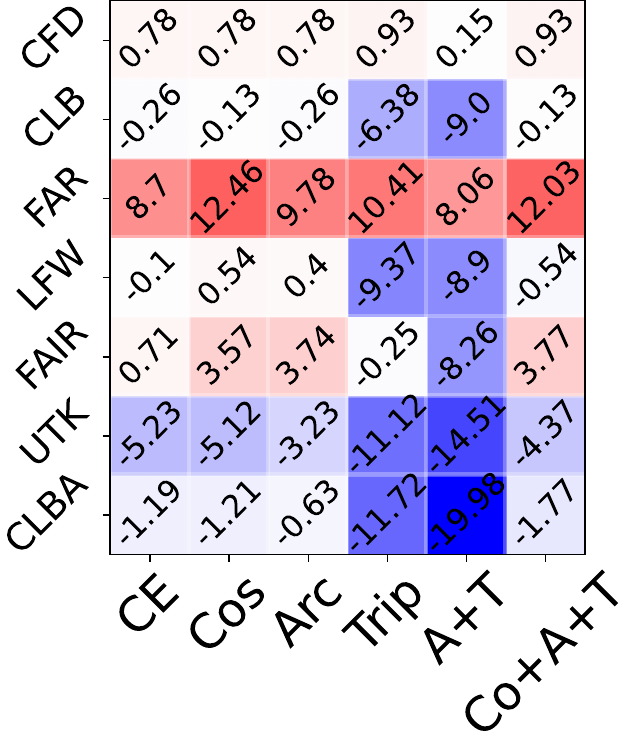}
        \caption{IBLIP}
	\end{subfigure}%
	\caption{Heatmaps indicating the extent and the direction of disparity between the two genders for different architectures across all datasets and loss functions. CelebSET is always disparate against males, and FARFace is always disparate against females, independent of all other factors, despite being balanced datasets. The color codes are as follows-- red indicates higher accuracy for males, and blue indicates higher accuracy for females. The intensity of the color signals the magnitude of the gender disparity. }
	\label{fig:heatmap-disp}
\end{figure*}

\begin{table}[!t]
    \centering
    \scriptsize
    \begin{tabular}{l| lc| lc}
        \toprule
        \multirow{2}{*}{Dataset} & \multicolumn{2}{c}{Accuracy ($\uparrow$)} & \multicolumn{2}{c}{Abs. Disparity ($\downarrow$)} \\\cmidrule{2-3}\cmidrule{4-5}
          & Ours & Vanilla & Ours & Vanilla \\ \midrule
          CFD & \textbf{98.33 \hfill(A)} & 97.29 & \textbf{0.98 \hfill(Cos)} & 4.57\\
          CLBST & 99.56 \hfill (Co+A+T) & \textbf{99.69} & 0.62 \hfill (Co+A+T) & \textbf{0.13}\\ 
          FAR & \textbf{93.07\hfill (Co+A+T)} & 91.68 & \textbf{12.78 \hfill(Co+A+T)} & 16.43 \\
          LFW & \textbf{97.76 \hfill (CE)} & 96.94 & \textbf{0.34 \hfill (Cos)} & 3.74\\
          UTK & 93.70\hfill (Co+A+T) & \textbf{94.73} & \textbf{0.03\hfill (T)} & 0.20\\
          CLBA & 90.14 \hfill (Co+A+T) & \textbf{90.35} & \textbf{0.79 \hfill (Co+A+T)} & 6.78\\
          \bottomrule
    \end{tabular}
    \caption{\blu{Accuracy \& disparity of the vanilla debiased model~\cite{karkkainen2021fairface} and the same model fine-tuned with our choice of loss functions. We only show the model+loss with highest accuracy \& lowest disparity.}}
    \label{tab:debiasing}
\end{table}

\begin{figure*}[t]
	\centering
	\begin{subfigure}{0.66\columnwidth}
        \centering
		\includegraphics[width= \textwidth, height=3.5cm, keepaspectratio]{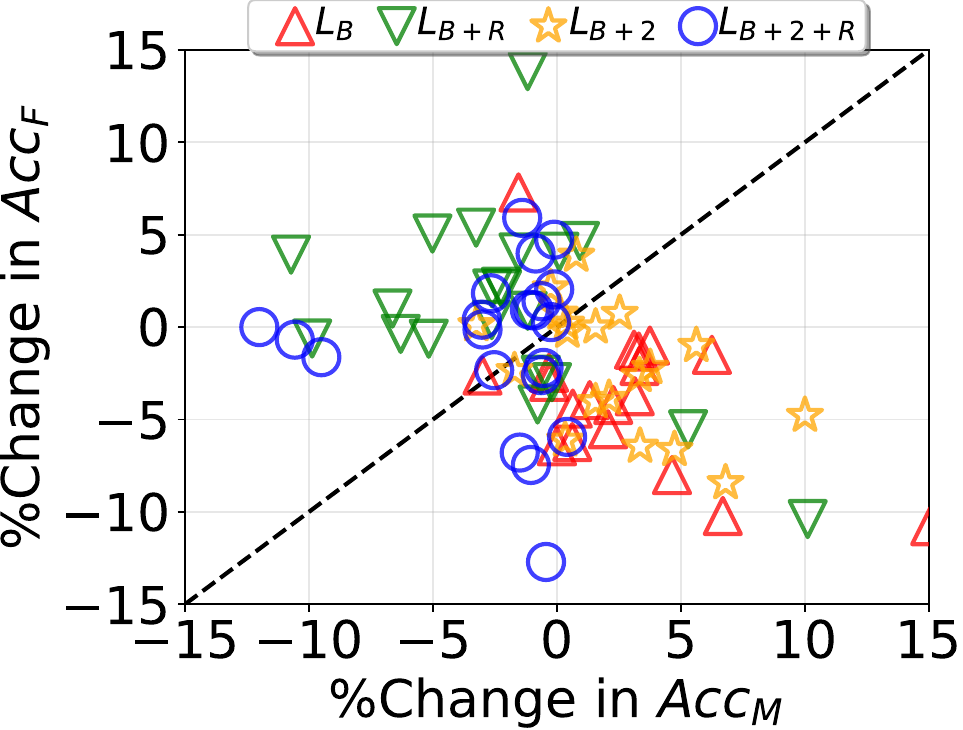}
        \caption{LBFC}
	\end{subfigure}%
	\begin{subfigure}{0.66\columnwidth}
        \centering
		\includegraphics[width= \textwidth, height=3.5cm, keepaspectratio]{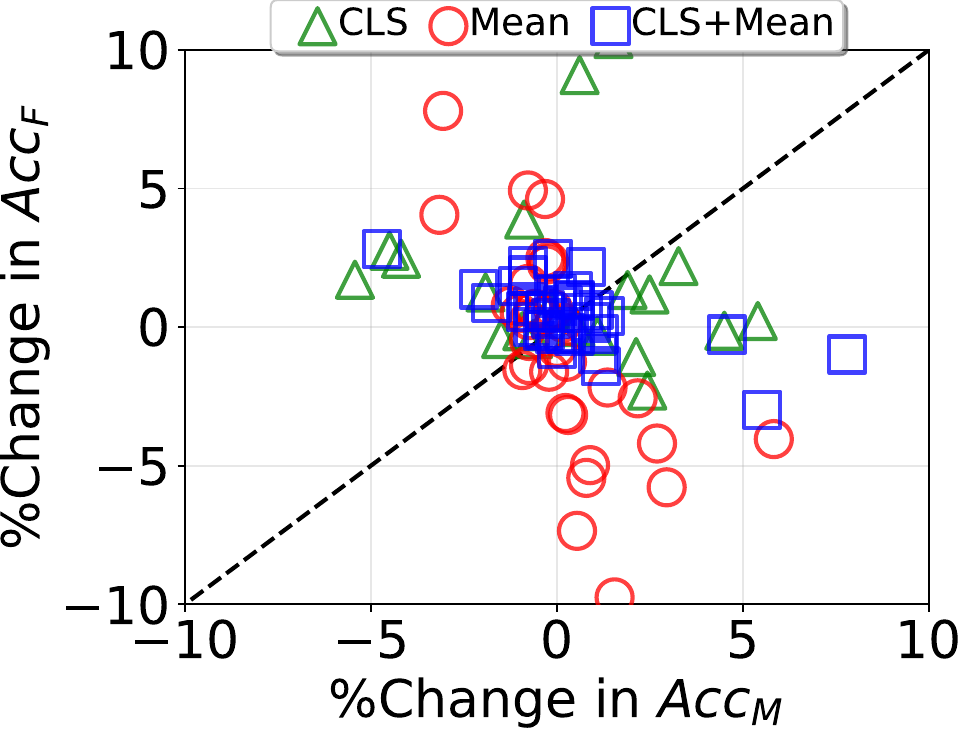}
        \caption{ViT-Face}
	\end{subfigure}%
 	\begin{subfigure}{0.66\columnwidth}
        \centering
		\includegraphics[width= \textwidth, height=3.5cm, keepaspectratio]{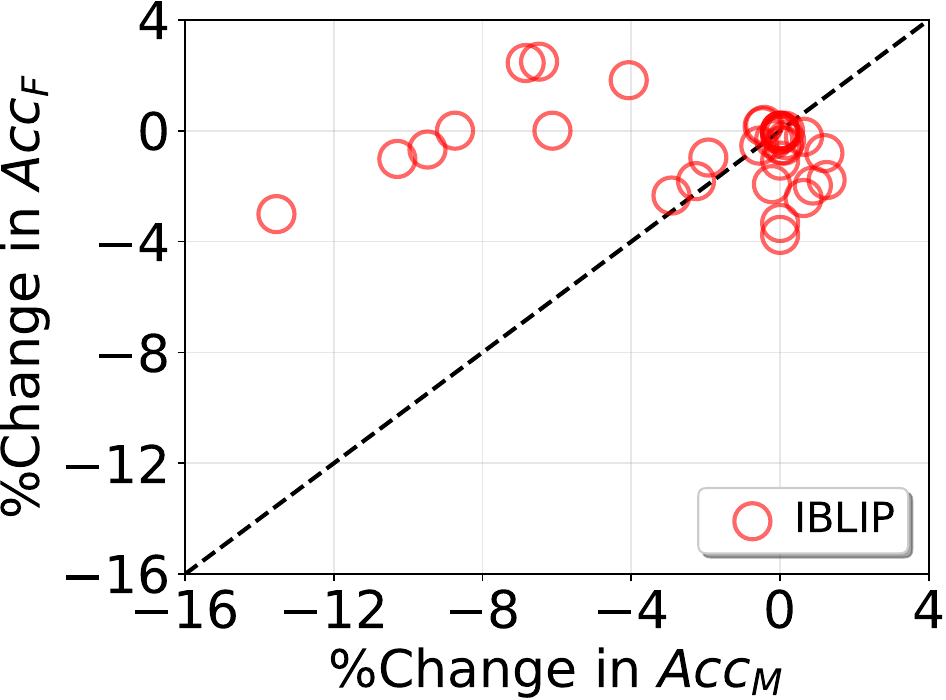}
        \caption{IBLIP}
	\end{subfigure}%
	\caption{Relative change in accuracy for males vs. females, for all losses when compared against only the CE loss, for all FRSs and datasets. The diagonal line indicates an equal relative change for both genders and points on either side imply a larger relative change for the respective gender. The effect of tinkering with the models has an impact on the relative change in accuracy for both genders.} 
	\label{fig:scatter-male-female}
\end{figure*}

\subsection{\blu{Direction of Disparity (RQ2)}}
Disparity in an FRS's performance, especially against social minorities, leads to denial of services~\cite{bestmirror2020,dunnepassport2019} and unfair treatment. Multiple studies in literature show that FRSs are heavily biased against females. In Figure~\ref{fig:heatmap-disp}, we look at heatmaps of disparity which reflect the group that has a higher accuracy-- red indicates males have a higher accuracy than females, and blue indicates vice versa; the intensity of each color reflects the extent of this disparity.

\subsubsection{Results for LibfaceID}
The heatmaps for the LibfaceID model generalize across architectural changes. From Figures~\ref{fig:heatmap-disp}a--d, we see (a)~All datasets report disparity, independent of the choice of architecture and loss function. Thus, the model is biased throughout and has a subpar performance in reducing disparity. \blu{(b)~From Figs.~\ref{fig:scatter-acc-disp}a--d, we already know FARFace and CelebA have the highest disparity. From the heatmaps, we see that this disparity in FARFace is \textit{always} against females, as previously observed~\cite{jaiswal2024breaking} and in CelebA is primarily against males}. UTKFace has the lowest average disparity. (c)~The choice of loss function impacts the disparity. Triplet loss has the highest disparity across datasets, independent of the architecture \blu{(except CelebA, where CE is the primary contributor to the highest disparity)}. (d)~Interestingly, we see that even though both CelebSET and FARFace are balanced datasets, their disparities are in exact opposite directions. \blu{CelebA, another dataset composed of celebrity faces, also reports disparities primarily against males.}
This shows that the choice of dataset is non-trivial, especially for evaluating FRSs, as it can lead to conflicting inferences.

\subsubsection{Results for ViT}
In Figures~\ref{fig:heatmap-disp}e--g, the heatmaps give a completely different picture as compared to LibfaceID. We see that (a)~The average disparities reduce significantly compared to the simpler CNN model.
(b)~In contrast to the CNN model, CelebSET and Fairface now report the lowest disparity instead of UTKFace. \blu{Using only the CLS embeddings leads to a bias against males, whereas using the mean of the patch embeddings reverts the bias to be against females (except the CelebSET \& CelebA dataset)}. As noted previously, FARFace has the highest disparity against females. \blu{(c)~The only observable pattern for the loss functions is triplet loss always reporting disparity against males in Fig.~\ref{fig:heatmap-disp}e (except for LFW) and against females in Figs.~\ref{fig:heatmap-disp}f-g.}
(d)~Here as well, both CelebSET and FARFace report biases in opposite directions, especially when all embeddings are used -- Fig.~\ref{fig:heatmap-disp}g.

\subsubsection{Results for InstructBLIP}
Finally, we take a look at the heatmap for the VLM in Fig.~\ref{fig:heatmap-disp}h. (a)~We observe the lowest average disparity for this model. (b)~The lowest disparities are observed for both CFD and CelebSET and the highest for FARFace and UTKFace. (c)~Similar to ViT, there is no generalizable pattern amongst the loss functions. \blu{(d)~Instead of CelebSET, the highest disparity against males is reported for CelebA (ArcFace + triplet loss)}. 
Interestingly, the average disparity against females is always higher than for males.

\subsubsection{Major takeaways}
We now list the major takeaways from the results in Figure~\ref{fig:heatmap-disp}.

\noindent $\bullet$ \textit{Model size determines disparity--} As we increase the model complexity (parameters, layers, etc.), the disparity reduces. Thus, larger models may be the key to reducing disparity.

\noindent $\bullet$ \textit{Choice of dataset determines perceived disparity--} Some datasets report disparity against females, whereas others report disparity against males, irrespective of the architecture and loss function. Thus, if the test sample changes, a model's direction and intensity of bias can become the opposite, irrespective of the learning involved. This is an important outcome not previously covered in audit studies, where most test datasets have similar distributions. Thus, in-the-wild out-of-distribution test points can make an FRS behave in a completely orthogonal way.

\noindent $\bullet$ \textit{Model complexity overpowers loss function--} In simpler CNN models, the loss function plays an important role, whereas in the complex transformer models, the architecture that already has an enormous advantage of the pre-training stage lends more weightage than the loss function in determining the disparity. 

Thus, we see that the answer to RQ2. is not monolithic; instead, the direction and intensity of disparity are heavily dependent on the choice of dataset and architecture. 

\subsection{\blu{Change in Male and Female Accuracies (RQ3)}}

\begin{table*}[!t]
    \centering
    \footnotesize
    \begin{tabular}{l c c c c | c c c c}
    \toprule
        \multirow{2}{*}{\textbf{Architecture}} & \multicolumn{4}{c|}{Highest accuracy} & \multicolumn{4}{c}{Lowest disparity} \\
        \cmidrule{2-9} 
        & \textbf{Loss} & \textbf{Data} & $D_M$ & $D_F$ & \textbf{Loss} & \textbf{Data} & $D_M$ & $D_F$\\
        \midrule
        LBFC$_B$ & A+T & CFD & 9.814 & \textbf{9.952} & A+T & UTKFace & 5.390 & \textbf{5.409}\\
        LBFC$_{B+R}$ & A+T & CFD & 8.185 & \textbf{8.330} & \blu{A} & \blu{LFW} & \blu{8.413} & \blu{\textbf{8.966}} \\
        LBFC$_{B+2}$ & A & CFD & 36.926 & \textbf{37.569} & A & UTKFace & \textbf{21.442} & 21.438\\
        LBFC$_{B+2+R}$ & A+T & CFD & 29.853 & \textbf{30.153} & A & UTKFace & \textbf{19.831} & 19.792\\
        \midrule
        ViT$_{CLS}$ & A & CFD & \textbf{25.086} & 24.903 & \blu{A} & \blu{LFW} & \blu{\textbf{26.301}} & \blu{24.969}\\
        \multirow{2}{*}{ViT$_{M}$} & A & \multirow{2}{*}{CelebSET} & \textbf{18.993} & 18.744 & \multirow{2}{*}{T} & \multirow{2}{*}{CelebSET} & \multirow{2}{*}{\textbf{15.270}} & \multirow{2}{*}{15.217}\\
        & Cos & & \textbf{25.381} & 24.798 & & & & \\
        ViT$_{CLS+M}$ & A+T & CFD & 21.618 & \textbf{21.693} & Co+A+T & CFD & 23.557 & \textbf{23.724}\\
        \midrule
        \multirow{2}{*}{IBLIP} & \multirow{2}{*}{Co+A+T} & \multirow{2}{*}{CelebSET} & \multirow{2}{*}{12.551} & \multirow{2}{*}{\textbf{12.635}} & Co+A+T & \multirow{2}{*}{CelebSET} & 12.551 & \textbf{12.635}\\
        & & & & & Cos & & 82.551 & \textbf{82.964} \\
    \bottomrule
    \end{tabular}
    \caption{Average Euclidean distance between the anchor embeddings and embeddings obtained with other loss functions for both gender groups ($D_M$ is avg. extent of shift for males and $D_F$ is for females), for the combinations that result in the highest accuracy and lowest disparity, respectively. Maximum values are in bold. On average, the embeddings for females shift more than for males.}
    \label{tab:embedding_dist}
\end{table*}

We now take a look at the relative change in accuracy for males and females in every architecture for all loss functions when compared against only the CE loss.
As mentioned in the previous section, all other losses are added along with the CE loss. In Figure~\ref{fig:scatter-male-female}, we plot this relative change as a scatter plot for each model. A diagonal line indicates an equal relative change for both males and females, and points lying on either side of the diagonal imply a larger relative change for one of the genders. Next, let us call the embeddings obtained using the simple CE loss the anchor embeddings.
We also measure the average Euclidean distance between these anchor embeddings and the embeddings obtained when other types of losses are used for each dataset and architecture. This shows the average extent of the shift of the male ($D_M$) and female ($D_F$) representations from their respective anchor embeddings. In Table~\ref{tab:embedding_dist}, we present the Euclidean distance values for the combinations that result in the highest accuracy and the lowest disparity, respectively (all other Euclidean distance values are in the Appendix).

\subsubsection{Results for LibfaceID}
We study the relative change in the accuracies for males and females for the different architectures in Fig.~\ref{fig:scatter-male-female}a. 
(a)~The scatter points have a large spread, thus indicating a lack of cohesive trend overall. On closer inspection, we note that the architectures with the residual connections are primarily placed above the diagonal on the top left quadrant, whereas the vanilla architectures are placed below the diagonal on the bottom right quadrant. This implies that adding residual connections favours a reduction in accuracy for males relative to an increase in accuracy for females and vice versa for the vanilla networks. 
(b)~From Table~\ref{tab:embedding_dist}, we see that both the highest accuracy and lowest disparity are generated whenever the ArcFace loss is present in the model's objective. 
(c)~For the highest accuracy scenario, $D_F$ is always higher than $D_M$. This shows that the embeddings for females shift a larger distance away from the anchor embeddings in the representation space as compared to males. On the other hand, for the lowest disparity scenario, the results are mixed. In the case of the shallow CNN, $D_F$ is higher (i.e., female representations shift away from the anchor embeddings more), while in the case of the deep CNN, the $D_M$ is higher (i.e., male representations shift away from the anchor embeddings more). 

\subsubsection{Results for ViT}
Next, we look at the results for ViT in Fig.~\ref{fig:scatter-male-female}b. 
(a)~We notice that each architecture behaves in a different manner-- the model with only CLS embeddings is spread horizontally (relative change for accuracy of males is higher for negligible change in accuracy for females) on both sides of the diagonal; the model with only the mean embeddings is spread vertically on both sides of the diagonal, \blu{and the model which uses a concatenation of both types of embeddings has its' points primarily clustered around the centre (it is minimal, but an equivalent change in the accuracy for both gender groups). }
(b)~From Table~\ref{tab:embedding_dist}, we observe that for the high-accuracy scenario, the avg. $D_M$ is higher than $D_F$ for both ViT$_{CLS}$ and ViT$_{M}$. On the other hand, the opposite is true for ViT$_{CLS+M}$. 
\blu{For the lowest disparity scenario, we make a similar observation.}

\subsubsection{Results for InstructBLIP}
Finally, from Fig.~\ref{fig:scatter-male-female}c, we note that-- 
(a)~A majority ($> 50\%$) of the points are clustered around the 0 mark, whereas the rest are either in a narrow vertical band, in the bottom right quadrant near 0, or strewn horizontally in the top left quadrant. Thus, most data points imply minimal change in accuracy for both genders.  
(b)~From Table~\ref{tab:embedding_dist}, we see that the best performance (high accuracy, low disparity) is observed when the CosFace loss is involved. Also, the values of $D_F$ are always higher than $D_M$.

\subsubsection{Major takeaways}
The major takeaways from Figure~\ref{fig:scatter-male-female} and Table~\ref{tab:embedding_dist} are as follows--

\noindent $\bullet$ \textit{Change in accuracy of genders is determined by the choice of model and architecture--} Residual connections in LibfaceID and CLS embeddings in ViT seem to affect the accuracy of the males. On the other hand, the vanilla CNN setup for LibfaceID affects the female accuracy more. Thus, the effect of the tinkering of the models is often tied to the sensitive attributes that the dataset encompasses.

\noindent $\bullet$ \textit{Embeddings of females are more sensitive to model choices than males--} The embeddings of females shift a larger distance from the anchor embeddings than males on average, for both the highest accuracy as well as the lowest disparity scenarios. This shows that the embeddings generated from images of females are more sensitive to the model changes-- architecture and loss function and could be an avenue for reducing disparity.

\subsection{\blu{Explaining the Observations}}
\begin{figure}[!t]
    \centering
    \includegraphics[width=0.45\textwidth]{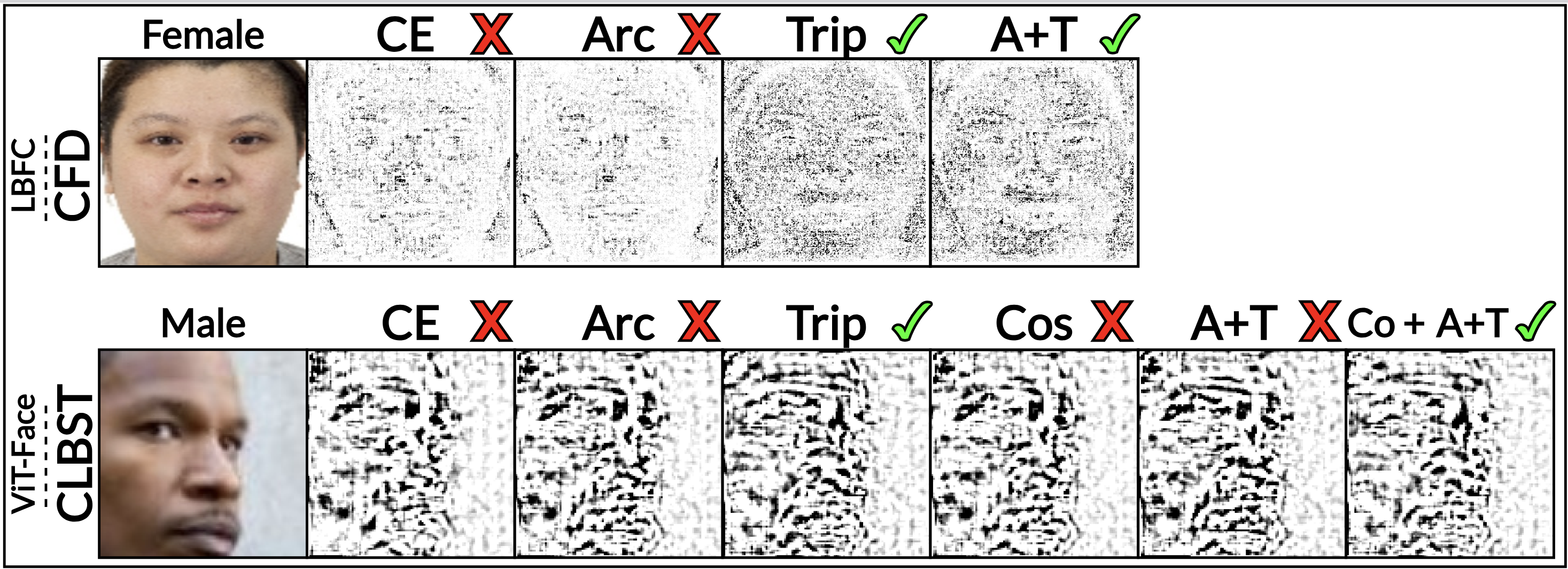}
    \caption{\blu{Explainability results on LBFC$_{B+2+R}$ (top row) and ViT$_{CLS+M}$ for sample images from the CFD \& CelebSET dataset with all loss functions using the method of integrated gradients~\cite{sundararajan2017axiomatic}. The original images, predictions and gradient attributions are presented.} }
    \label{fig:captum-cnn}
\end{figure}

\blu{We use the method of integrated gradients~\cite{sundararajan2017axiomatic} to infer the reason for the predictions for each of the models. In Figure~\ref{fig:captum-cnn}, we present the results for the explainability experiments on LBFC$_{B+2+R}$ (top row) and ViT$_{CLS+M}$ (bottom row) models with all loss functions for sample images from the CFD and CelebSET datasets respectively. We see how each model \& loss function elicits a starkly different gradient attribution. Whenever the CNN model predicts a person as male, the attributions are more ``localized'', and the focus is on the eyebrows, nose, jawline or facial hair, and whenever the model predicts a person as female, the focus is more ``globalized'', distributed and balanced with a focus on the upper facial contours. In the top row, the correct prediction is thus obtained for the T and A+T loss functions that enforce localized attention. In the ViT model, we notice patch-based grid-like attributions wherein the attention is more on the eyes and upper face region when predicting female and more uniform throughout the face when predicting male. In the bottom row, the correct prediction is thus obtained for the T and Co+A+T loss functions that enforce uniform attention.
While we present only a subset of results here due to paucity of space, the inferences from the other dataset-model-loss function combinations are very similar.}
\section{\blu{Discussion and Recommendations}}
\label{sec:discussion}
We now discuss the high-level learnings from our extensive audit as well as the recommendations for future auditors, developers and users of such FRSs.

\subsection{Summary}
In this work, we attempted to understand how the three major components of an FRS impact the model's performance when observing two opposing objectives -- accuracy and disparity, for the task of gender prediction. To address RQ1, we studied the accuracy vs disparity trade-off amongst the different architectures, loss functions and datasets from Figure~\ref{fig:scatter-acc-disp}. We trivially observed that with increasing model complexity, the accuracy increases, and the disparity reduces as models with deeper layers learn more salient discriminative features in the face images. Next, we stress on the importance of choosing diverse datasets -- CFD~\cite{ma2015chicago}, the most standard face dataset in our benchmark, consistently reports the highest accuracies (especially on the shallow CNN). CelebSET~\cite{raji2020saving}, a dataset of Hollywood celebrity faces, reports the highest accuracies on both ViT and InstructBLIP. We hypothesize that this is possibly due to data leakage from the pre-training stage. Finally, \blu{CelebA \&} FARFace~\cite{jaiswal2024breaking}, a newly released dataset with significantly more images from the Global South who have different skin tones and facial features, prove to be the most adversarial, reporting high disparity. Hence, the choice of dataset may propagate a false sense of confidence or fear in the model's abilities and stakeholders need to better engage with the datasets in order to understand the FRS's true performance. \blu{This is especially true for social media platforms as they are deployed worldwide and disparities can impact millions of users.}
We observed a strong tendency of clustering for each face dataset, with regards to the performance metrics, that is independent of the model and loss function under consideration. 

Next, to address RQ2, we studied the heatmaps in Figure~\ref{fig:heatmap-disp} for the extent and direction of disparity in each model for all architectures, loss functions and datasets. We discovered that the model size and dataset determine not only the extent of disparity but also the direction of disparity. For example, both CelebSET and FARFace are gender-balanced datasets of ``in-the-wild'' photos of public-facing individuals (Hollywood celebrities and cricketers, respectively), yet CelebSET consistently reports disparity against males and FARFace reports disparities against females. This again confirms that the choice of dataset has a huge impact on \textit{the kind of bias a model is perceived to have}. We also trivially observed that in complex models, the loss functions play a less important role in determining the extent and direction of disparity. 

In Figure~\ref{fig:scatter-male-female} and Table~\ref{tab:embedding_dist}, we attempted to quantify the relationship between the change in accuracy for the two gender groups to address RQ3. Our results show that the combination of architecture and loss function determines which gender group reports a higher accuracy and to what extent for all datasets. For example, ArcFace loss always gives the highest accuracy and lowest disparity, independent of the architecture; in the two most optimal models-- VIT$_{CLS+M}$ and InstructBLIP, the embeddings for females shift more than males. This shows that existing datasets have a lot of variety in what constitutes a ``female face'', and there is no general pattern for the same as for males. We hypothesize here that this is the leading cause of bias in FRSs, and ML developers need to design better models that can find more generalizable features, especially for female faces, to address the problem of disparity.

\subsection{\blu{Recommendations for the Community}}
We perform an in-depth investigation into the relationship between model architecture, loss function and data and its ensuing impact on disparity for the task of gender prediction. Our experimental results show that all three components, individually as well as collectively, impact both accuracy and disparity. Any human-facing technology, especially one with the sensitive nature of deployment like an FRS, has multiple stakeholders -- developers, users and regulators, all of whom need to perform due diligence before the technology can be released to the general public. From our study, we make some recommendations for developers and users.
First, the model developers need to be cognizant of the various architectures and how they interact with each dataset (or the population where the model will be deployed)\footnote{\url{https://www.paravision.ai/news/addressing-critical-inclusion-questions-for-face-recognition-technology-buyers/}}. Similarly, the users need to be aware of the deployment scenario and which model suits their use case~\cite{cherepanova2023deepdive}. For example, a model to be deployed in the USA should be audited and designed with the CelebSET dataset as a template, whereas one that has to be deployed in India or the West Indies should use the FARFace dataset. We believe our study will provide the necessary awareness and blueprint that the community can benefit from and reduce the disparity and unfair outcomes that result from the bias in these models.

\noindent \blu{\textit{Recommendations for social media platforms}: Our results are especially useful for social media platforms that use FRSs for image tagging, advertising and scammer detection applications. We show in our experiments how each model performs differently over a given dataset and the associated disparities differ. Thus, social media platforms deployed across the world should take cognizance of this report and deploy more localized models that are trained on data reflective of the local geography to reduce instances of disparate performance. Further, various FRS companies, e.g., Clearview AI\footnote{\url{https://www.businessinsider.com/clearview-scraped-30-billion-images-facebook-police-facial-recogntion-database-2023-4?utm_source=chatgpt.com}} routinely use billions of datapoints scraped from social media platforms to build their database to be shared with law enforcement agencies thus putting everyone into a ``perpetual police line-up''. Biases in this type of technology would only exacerbate the risks of false arrests or detentions. To minimize such risks, the social media platforms thus need to be extra vigilant perhaps making amendments to their data scraping/download policies.} 

\subsection{\blu{Limitations of our Study}}
\blu{Similar to many other studies in this field, ours also has certain limitations, which we acknowledge and clarify here.}

\noindent\blu{\textbf{Definition of gender}: Historically, gender has been classified as binary-- masculine and feminine. In this model, gender is aligned with the sex assigned at birth. Such definitions often conform to legal, governmental, political, societal and even technological expectations, wherein having the lowest common denominator definition suits most use cases. Most government IDs only recognize the binary male or female gender identity, which technology like face recognition systems has to adhere to, especially in security applications like airport entry. This binary definition has been criticized by scholars of intersectionality, who maintain that such a structure maintains patriarchal and supremacist norms~\cite{scaptura2023systems}. We acknowledge and reiterate that gender is fluid and can include various groups like transgender, agender, intersex and other non-binary identifying individuals. Due to the reasons stated here, we are limited by the choice of only binary gender labels available in all existing datasets. Thus, predictions should be interpreted as the ``\textit{perceived}'' binary gender.}

\noindent \blu{\textbf{Dataset artifacts}: We understand that each face dataset has certain artifacts like representation of a narrow socio-economic demographic, age group and other factors, which may limit the generalizability of our findings. To circumvent this issue to an extent we have selected a multitude of datasets roughly covering a mix of both the well-studied and the newly introduced ones. Further construction of large-scale datasets encompassing different sociodemographic backgrounds seems imperative. This points to an interesting future direction of research.
}

\section{Conclusion}
In this study, we investigate the role of datasets, model architectures and loss functions in shaping the accuracy-disparity landscape in FRSs for the task of gender prediction. 
We use three different models-- LibfaceID (a CNN architecture), ViT-Face (a vision transformer architecture) and InstructBLIP (a vision-language model architecture), \blu{seven} diverse face datasets -- CFD, CelebSET, FARFace, \blu{LFW}, UTKFace, Fairface and \blu{CelebA}, and four loss functions-- Cross Entropy, Triplet, ArcFace and CosFace loss functions. We modify the FRSs into multiple variants to create ten models that are then trained and fine-tuned with a combination of the loss functions mentioned here, resulting in \blu{266} evaluation configurations.

Our large-scale study shows that the choice of the loss function and evaluation dataset is very important to understand the true bias of the FRS. Highly standard datasets like CFD result in high accuracy and low disparity, whereas adversarial datasets like FARFace \blu{\& CelebA} expose the true shortcomings of the FRS. Triplet loss, used in previous studies to improve the performance of FRSs does not work as well when used on different architectures and datasets. Similarly, model size and dataset combine to determine which gender performs better, with two ``in-the-wild'' datasets with popular individuals reporting disparities in exactly opposite directions. Finally, we observe that the architecture and loss function also combine to determine which gender group has a higher accuracy and to what extent for all datasets. Our most interesting observation is the fact that existing datasets have a lot of variety for ``female faces'' but not so much for male faces, which could be a leading cause of bias in FRSs.

\subsection{\blu{Future Work}}
We plan to extend this work to larger and more diverse vision architectures and a more diverse choice of loss functions and datasets. We also plan to use our findings to devise FRS debiasing algorithms that can improve the performance for the genders reporting lower accuracy and bring more parity in the model's performance.
\bibliography{main}
\newpage
\section{Ethics Statement}
\label{sec:ethics}
Our in-depth study shows that FRSs have disparity irrespective of the dataset, choice of architecture or loss functions. Thus, existing algorithms and datasets are still not apt for large-scale deployment in society. Our study attempts to shed light on the reason for this disparity, which model developers and users can take cognizance of for their own deployment use cases. We acknowledge that gender is a spectrum but limit ourselves to binary gender prediction due to the gender labels available in the benchmark datasets. We plan to share model cards~\cite{mitchell2019model} of all our FRS models upon acceptance aligning with Responsible AI practices.
\newpage

\section{Appendix}
\label{sec:appx}

\subsection{Additional details on experimental design}
\subsubsection{Architectures and loss functions for LibfaceID}
In addition to the models described in the main draft, we create two more models with weighted residual connections. The number of layers and connections are the same as described in Fig.~\ref{fig:libfaceid_models}. The weight of a residual connection determines how much information is passed on to the next layer through this connection, calculated using the given formula--

\centerline{$\alpha \times O_{L-1} + (1 - \alpha) \times O_L$} where $O_L$ is the output for layer $L$ and $\alpha$ is a heuristic. We perform a grid search using Optuna~\cite{akiba2019optuna} to calculate the value of $\alpha$ as 0.3.

For all models where we use ArcFace and Triplet loss along with Cross-Entropy loss, the weights are-- $CE + \alpha T + \beta A$, where $(\alpha,\beta) = (0.7,0.1)$.

\subsubsection{Loss functions for vision transformer}
For ViT$_M$ and ViT$_{CLS+M}$, we use CosFace, ArcFace and Triplet loss with Cross-Entropy loss in various combinations, whose weights are determined as follows-- {(a)}~$CE + \alpha T + \beta A$, where $(\alpha,\beta) = (0.6,0.6)$ for ViT$_M$ and $(\alpha,\beta) = (0.8,0.9)$ for ViT$_{CLS+M}$, and {(b)}~$CE + \alpha T + \beta A + \gamma Cos$, where $(\alpha,\beta, \gamma) = (0.9,0.4,1)$ for ViT$_M$ and $(\alpha,\beta, \gamma) = (0.4,1,0.8)$ for ViT$_{CLS+M}$.

\subsubsection{Loss functions for InstructBLIP}
For InstructBLIP, the combination of CosFace, ArcFace and Triplet loss with Cross-Entropy loss are weighted in the following manner-- (a)~$CE + \alpha T + \beta A$, where $(\alpha,\beta) = (0.9,0.4)$, {(b)}~$CE + \alpha T + \beta A + \gamma Cos$, where $(\alpha,\beta,\gamma) = (0.8,1,0.5)$.

\subsection{Hyperparameters}
In Table~\ref{tab:hyperparameters}, we present the key hyperparameters -- the learning rate (LR), the optimizer and the number of epochs -- used in our experiments. 
\begin{table}[!ht]
    \centering
    \footnotesize
    \begin{tabular}{l| c| c| c}
    \toprule
        Model & LR & Optimizer & \#Epochs\\
        \midrule
        LBFC$_B$ & \blu{$9 \times 10^{-3}$} & \multirow{6}{*}{SGD} & \multirow{10}{*}{100} \\ 
        LBFC$_{B+R}$ & \blu{$2 \times 10^{-3}$} & ~ & ~ \\ 
        LBFC$_{B+R\alpha}$ & \blu{$9 \times 10^{-3}$} & ~ & ~ \\ 
        LBFC$_{B+2}$ & \blu{$1 \times 10^{-3}$} & ~ & ~ \\ 
        LBFC$_{B+2+R}$ & \blu{$9 \times 10^{-3}$} & ~ & ~ \\ 
        LBFC$_{B+2+R\alpha}$ & \blu{$6 \times 10^{-4}$} & ~ & ~ \\ \cline{1-3}
        ViT$_{CLS}$ & \blu{$7 \times 10^{-4}$} & \multirow{4}{*}{AdamW} & ~ \\ 
        ViT$_{M}$ & \blu{$3 \times 10^{-5}$} &  & ~ \\ 
        ViT$_{CLS+M}$ & \blu{$1 \times 10^{-4}$} &  & ~ \\ \cline{1-2}
        IBLIP & \blu{$2 \times 10^{-3}$} &  & ~ \\
    \bottomrule
    \end{tabular}
    \caption{Training and fine-tuning hyperparameters.}
    \label{tab:hyperparameters}
\end{table}

\subsection{Results on Adience}
\subsubsection{Experimental Details}
We split the Adience dataset into train, validation, and test sets in an 80:10:10 ratio. Thus, we have 810 males and 935 females in the test set for Adience. We present the results for this set as follows.

We train/fine-tune our FRS models on an Ubuntu 18.04 LTS Intel(R) Xeon(R) Gold 6126 CPU server with NVIDIA Tesla P100 GPU (CUDA v12.2) $\times$ 2, 128 GB RAM and 48 cores. The model hyperparameters are present in Table~\ref{tab:hyperparameters}, chosen using Optuna.

\begin{figure}[!t]
	\centering
	\begin{subfigure}{0.66\columnwidth}
        \centering
		\includegraphics[width= \textwidth, height=3.5cm, keepaspectratio]{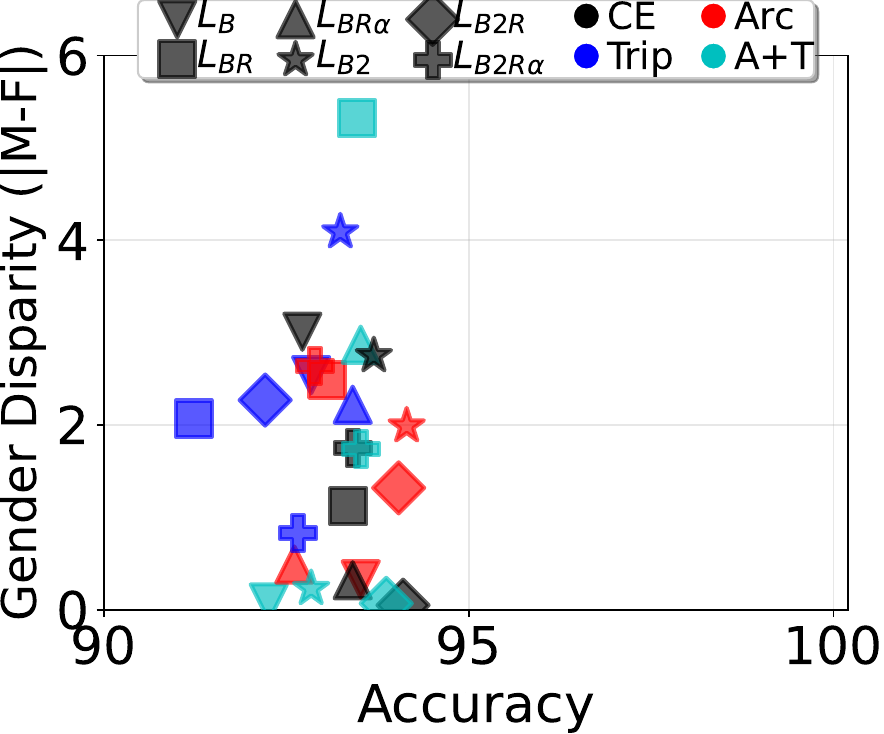}
        \caption{LBFC}
	\end{subfigure}\\
	\begin{subfigure}{0.66\columnwidth}
        \centering
		\includegraphics[width= \textwidth, height=3.5cm, keepaspectratio]{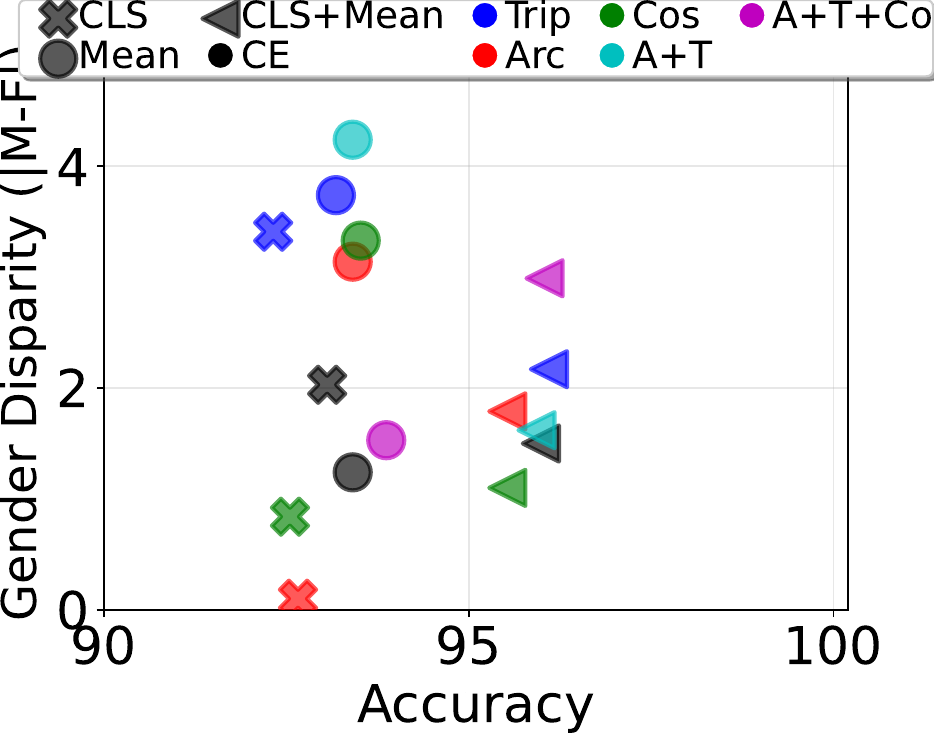}
        \caption{ViT}
	\end{subfigure}\\
 	\begin{subfigure}{0.66\columnwidth}
        \centering
		\includegraphics[width= \textwidth, height=3.5cm, keepaspectratio]{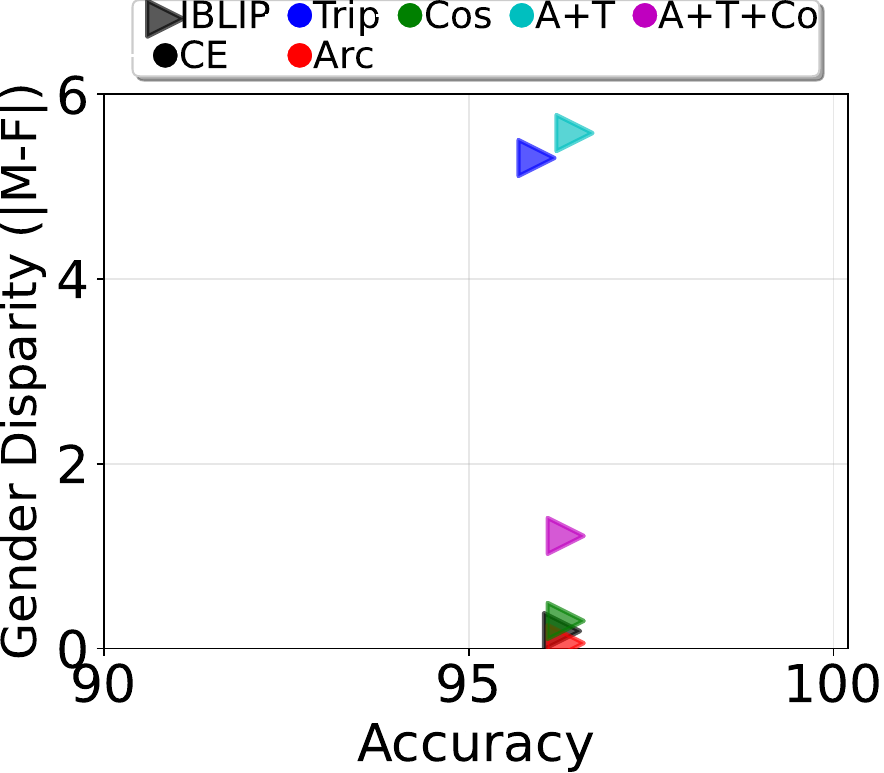}
        \caption{IBLIP}
	\end{subfigure}%
	\caption{Accuracy vs. absolute gender disparity for the Adience test set, for each type of model backbone, independent of the loss functions. The accuracy increases with increasing model complexity. The absolute disparity remains consistently upper bounded at 6\%.} 
	\label{fig:ad-sctt}
\end{figure}

\subsubsection{Accuracy vs disparity}
From Fig.~\ref{fig:ad-sctt}, we can see that for all FRSs, the scatter points lie in a vertical line, indicating a relatively stable accuracy but diverse absolute disparity values. We also note that with increasing model complexity, the accuracy value increases as well. Interestingly, the disparity remains consistent, upper bounded at 6\%. For ViT (Fig.~\ref{fig:ad-sctt}b), we also note that when the type of embedding changes from trivial to more detailed, the model accuracy increases.

\begin{figure}[!t]
	\centering
	\includegraphics[width= \textwidth, height=3.5cm, keepaspectratio]{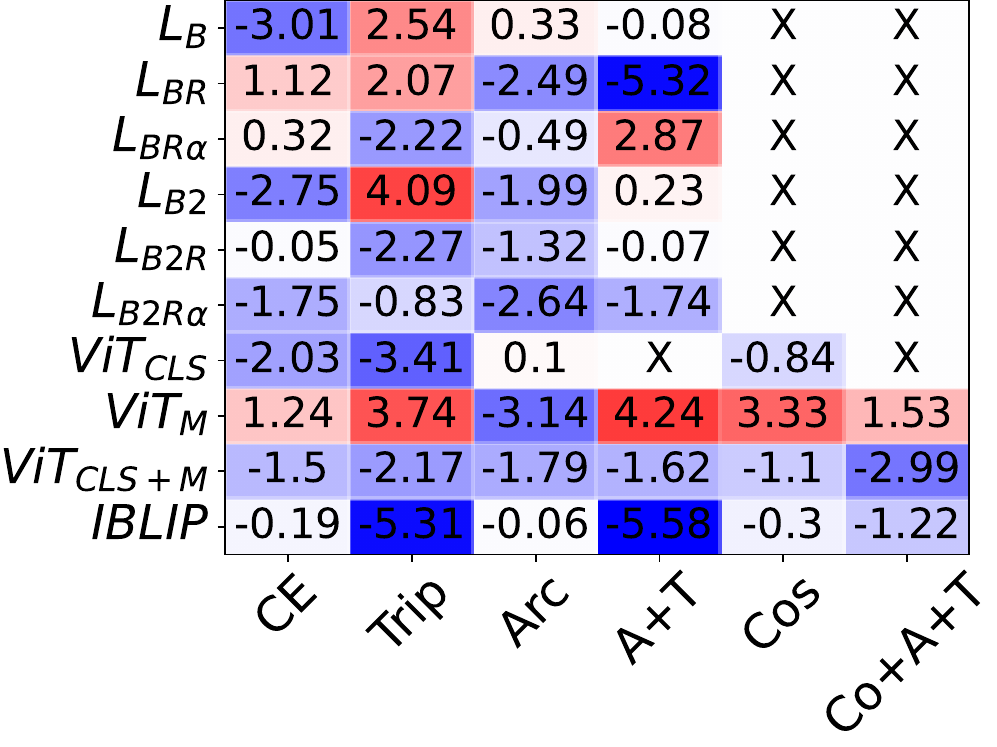}
	\caption{Heatmap for each type of FRS and loss functions tested on Adience. This plot shows the direction of disparity-- against females (+ve) or against males (-ve) across all the models and losses.} 
	\label{fig:ad-hmp}
\end{figure}

\subsubsection{Direction of disparity}
In Fig.~\ref{fig:ad-hmp}, we look at the heatmaps of disparity for the different loss functions and FRSs. The color indicates the direction of disparity, with red implying higher accuracy for males and blue implying higher accuracy for females. Since the LibfaceID model is trained from scratch on the Adience dataset, which has more female images, the test accuracy for females is naturally higher (as evident from the first 6 rows of Fig.~\ref{fig:ad-hmp}, especially for models not using Triplet loss). For ViT, we note that only ViT$_M$ reports disparity against females. Finally, InstructBLIP reports disparities against males as well.

\begin{figure}[!t]
	\centering
	\includegraphics[width= \textwidth, height=3cm, keepaspectratio]{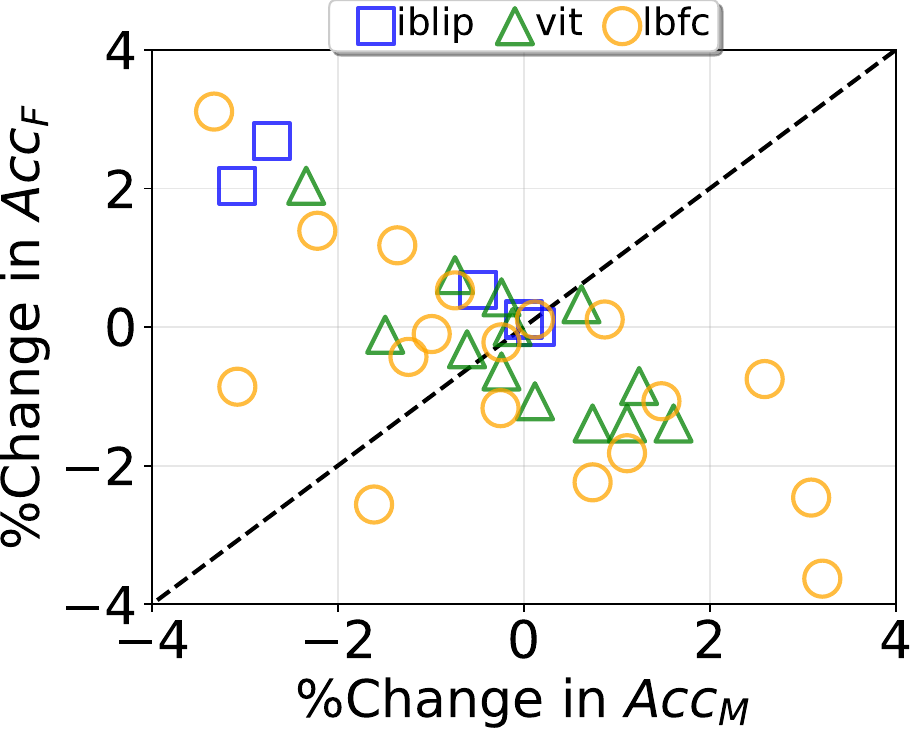}
	\caption{Change in male and female accuracies for each FRS tested on Adience.} 
	\label{fig:ad-mf}
\end{figure}

\subsubsection{Change in male and female accuracies}
In Figure~\ref{fig:ad-mf}, we observe the change in male and female accuracies for each FRS-architecture-loss combination on the Adience test set. For all FRSs, we note that the scatter points are distributed along the cross-diagonal, implying that there is an inverse relationship between the change in accuracies for males and females. For InstructBLIP, we note that all scatter points are above the diagonal-- the change in accuracy for males is always negative, and for females is always positive.

\begin{figure}[!t]
	\centering
	\begin{subfigure}{0.66\columnwidth}
        \centering
		\includegraphics[width= \textwidth, height=3.5cm, keepaspectratio]{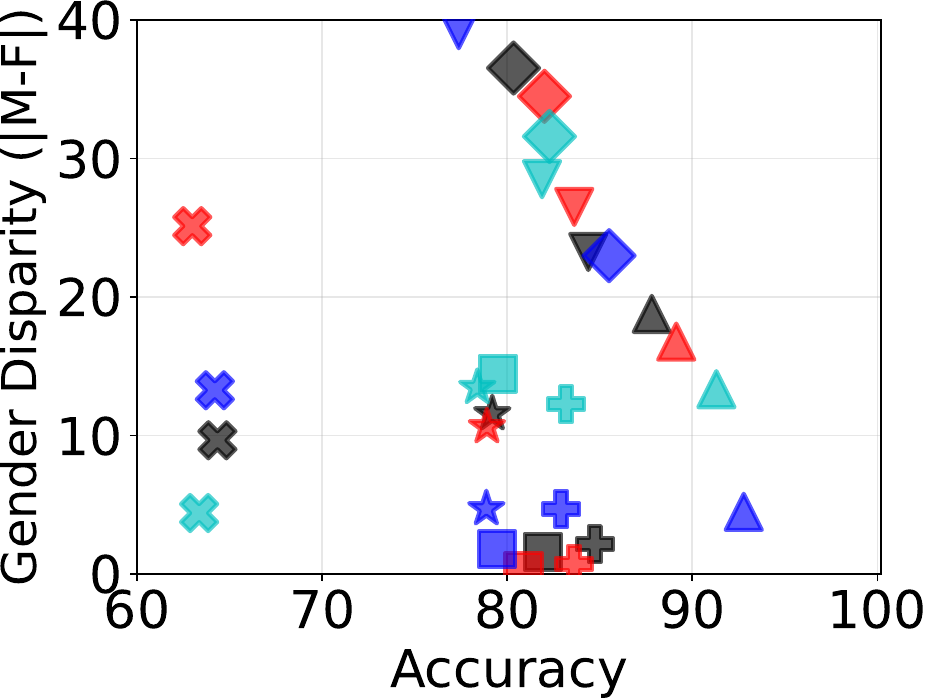}
        \caption{LBFC$_{B+R\alpha}$}
	\end{subfigure} \\
	\begin{subfigure}{0.66\columnwidth}
        \centering
		\includegraphics[width= \textwidth, height=3.5cm, keepaspectratio]{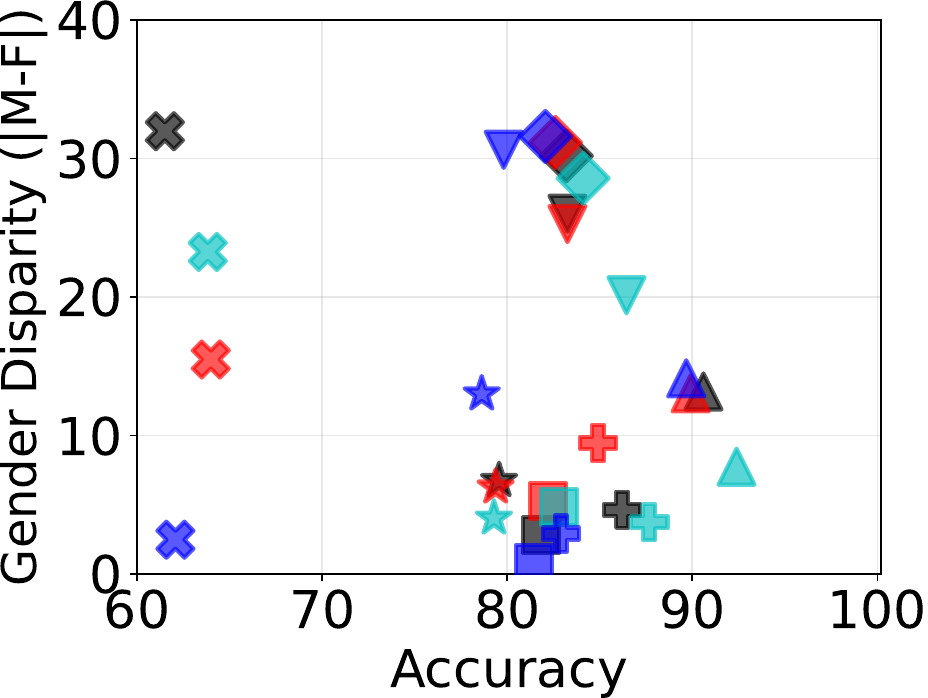}
        \caption{LBFC$_{B+2+R\alpha}$}
	\end{subfigure}%
        \\
    \begin{subfigure}{0.52\textwidth}
        \centering
        \includegraphics[width= 0.75\textwidth, height=2.5cm, keepaspectratio]{plots/scatter-acc-disp/Legend-Fig3_28-12.pdf}
    \end{subfigure}
 	\caption{Accuracy vs absolute gender disparity for the LibfaceID FRS with weighted residual connections.} 
	\label{fig:lbfc-alpha-acc-disp}
\end{figure}

\begin{figure}[!t]
	\centering 
        \begin{subfigure}{0.48\columnwidth}
        \centering
		\includegraphics[width= \textwidth, height=3.5cm, keepaspectratio]{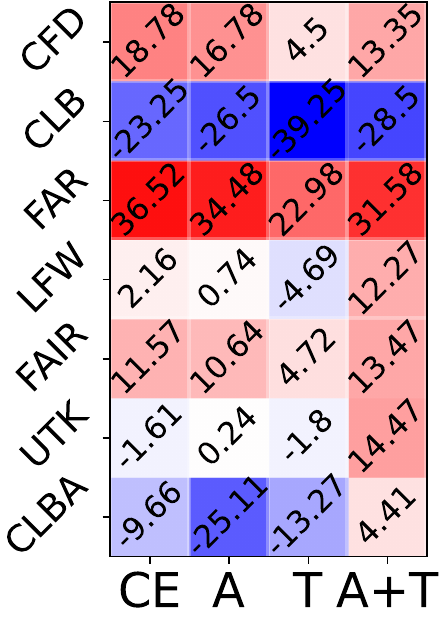}
        \caption{LBFC$_{B+R\alpha}$}
	\end{subfigure}~
        \begin{subfigure}{0.48\columnwidth}
        \centering
		\includegraphics[width= \textwidth, height=3.5cm, keepaspectratio]{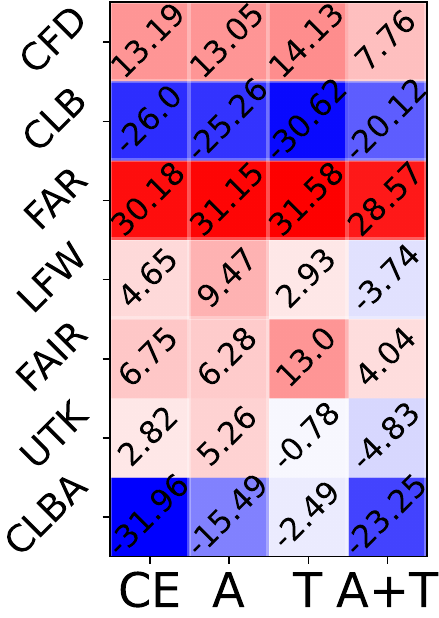}
        \caption{LBFC$_{B+2+R\alpha}$}
	\end{subfigure}%
	\caption{Direction of disparity for all datasets and all loss functions on the LibfaceID FRS with weighted residual connections.} 
	\label{fig:lbfc-alpha-hm}
\end{figure}

\begin{figure}[!t]
	\centering
	\includegraphics[width= \textwidth, height=3.5cm, keepaspectratio]{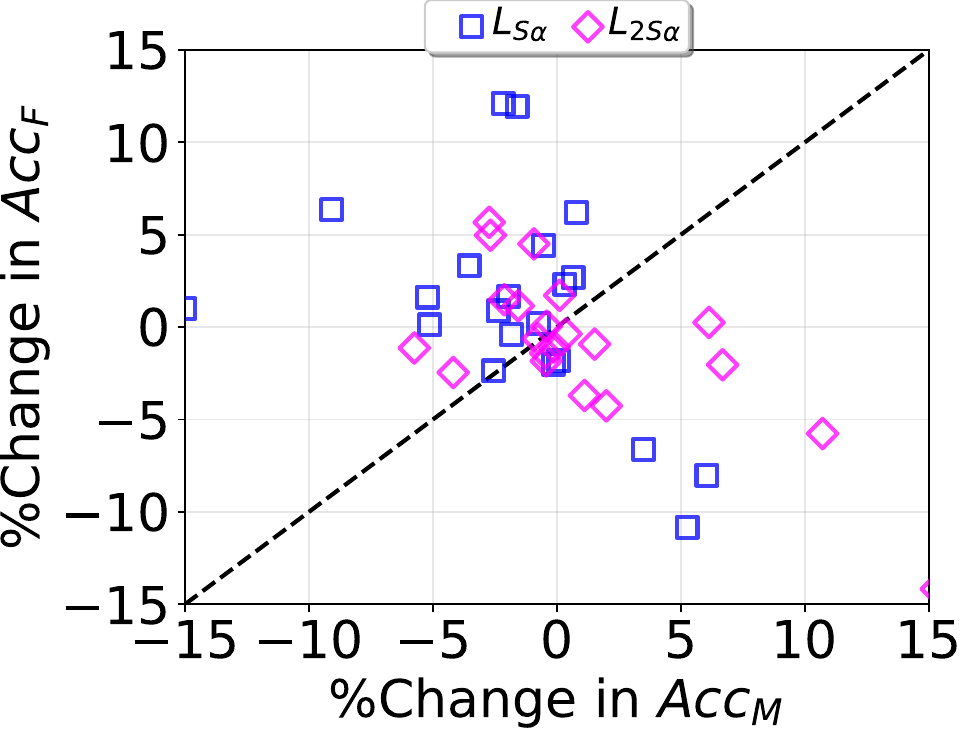}
	\caption{Change in male and female accuracies for the LibfaceID FRS with weighted residual connections.} 
	\label{fig:lbfc-alpha-mf}
\end{figure}

\subsection{Results on LibfaceID with weighted residual connections}
We now look at the experimental results on the LibfaceID FRS with weighted residual connections in Figures~\ref{fig:lbfc-alpha-acc-disp},~\ref{fig:lbfc-alpha-hm} and~\ref{fig:lbfc-alpha-mf}. 
The results are similar to those observed for other LibfaceID models (discussed in the main draft). We discuss each result as follows-- 

\begin{compactitem}
    \item In Fig.~\ref{fig:lbfc-alpha-acc-disp}, we can see that FARFace always reports the highest disparity, whereas UTKFace reports the lowest. Next, CFD reports the highest accuracy and \blu{CelebA} the lowest. We also notice a strong clustering in the results for each dataset with regard to one of the objectives.
    \item In the heatmaps in Figure~\ref{fig:lbfc-alpha-hm}, we see that CelebSET and FARFace always report the highest disparity across architecture and loss function variations. Interestingly, CelebSET and FARFace are both balanced datasets, however their disparities are in exact opposite directions. UTKFace always reports the lowest disparity among all datasets (Figure~\ref{fig:lbfc-alpha-acc-disp}). 
    \item In Figure~\ref{fig:lbfc-alpha-mf}, we look at the change in male and female accuracies. Here, we can observe that the points lie primarily on or above the diagonal, indicating that either the accuracies for males and females do not change much, and if they do, then the accuracy for males increases less often than the accuracy for females increases.
\end{compactitem}

\subsection{Change in embeddings for males and females}
In Tables~\ref{tab:eucl_lbfc_b}--\ref{tab:eucl_iblip}, we look at the change in the position of the embeddings generated for males and females for each FRS in the different architecture, loss function and dataset combination. The metric we use for calculating this change is the Euclidean distance between the embeddings generated with only the Cross-Entropy loss and other loss function combinations.
The key takeaways are as follows --

\begin{itemize}
    \item Among all the datasets, CFD always has a higher Euclidean distance for females than males, whereas FARFace always has a higher Euclidean distance for males instead of females. This indicates the sensitivity of each gender group in individual datasets to the model architecture and loss function. FARFace reports the largest difference between the distance measures for males and females. 

    \item The results for all LibfaceID models (Tables~\ref{tab:eucl_lbfc_b}--\ref{tab:eucl_lbfc_b2ra}) report a higher Euclidean distance value for males than females. On adding residual connections to the architecture (Tables~\ref{tab:eucl_lbfc_br} and \ref{tab:eucl_lbfc_b2r}), the number of datasets following this trend increases, as opposed to non-residual networks.

    \item From Table~\ref{tab:eucl_vit_cls}, we see that the change in embedding positions for males and females in ViT$_{CLS}$ is closer than observed for LibfaceID.
    Tables~\ref{tab:eucl_vit_m} and \ref{tab:eucl_vit_cat} show that across different loss functions, the behaviour of the datasets remains the same. For both models, Adience, CelebSET and FARFace always report higher Euclidean distance measures for males than females, indicating that male images observe a larger shift in the embedding space than females. 

    \item In Table~\ref{tab:eucl_iblip}, we observe the results for InstructBLIP.  
    The trend in the change of embedding position is constant for males and females across all the loss functions. Interestingly, upon applying the CosFace loss to the FRS, the Euclidean distance is observed to be the highest. This shows that the CosFace loss changes the positions of the embeddings the most out of all loss functions. 
\end{itemize}

\begin{table}[!t]
    \centering
    \footnotesize
    \begin{tabular}{c c c c}
    \toprule
        \textbf{Loss} & \textbf{Data} & $D_M$ & $D_F$ \\
        \midrule
        \multirow{6}{*}{T} & Adience & 6.065 & \textbf{6.310} \\
        & CelebSET & \textbf{6.802} & 6.663 \\
        & CFD & 9.816 & \textbf{9.964} \\
        & FairFace & \textbf{5.414} & 5.251 \\
        & FARFace & \textbf{9.911} & 5.335 \\
        & UTKFace & 5.417 & \textbf{5.436} \\
        & \blu{LFW} & \blu{5.344} & \blu{\textbf{5.627}} \\
        & \blu{CelebA} & \blu{4.232} & \blu{\textbf{4.859}} \\
        \midrule
        \multirow{6}{*}{A} & Adience & \textbf{8.621} & 8.451 \\
        & CelebSET & \textbf{9.225} & 9.023 \\
        & CFD & 13.471 & \textbf{13.640} \\
        & FairFace & \textbf{7.614} & 7.169 \\
        & FARFace & \textbf{13.755} & 7.644 \\
        & UTKFace & 7.452 & \textbf{7.469} \\
        & \blu{LFW} & \blu{\textbf{7.512}} & \blu{7.458} \\
        & \blu{CelebA} & \blu{6.304} & \blu{\textbf{6.739}} \\
        \midrule
        \multirow{6}{*}{A+T} & Adience & 6.070 & \textbf{6.195} \\
        & CelebSET & \textbf{6.769} & 6.631 \\
        & CFD & 9.814 & \textbf{9.953} \\
        & FairFace & \textbf{5.397} & 5.172 \\
        & FARFace & \textbf{9.970} & 5.264 \\
        & UTKFace & 5.390 & \textbf{5.410} \\
        & \blu{LFW} & \blu{5.369} & \blu{\textbf{5.539}} \\
        & \blu{CelebA} & \blu{4.194} & \blu{\textbf{4.829}} \\
    \bottomrule
    \end{tabular}
    \caption{Average Euclidean distance for LBFC$_B$.}
    \label{tab:eucl_lbfc_b}
\end{table}

\begin{table}[!t]
    \centering
    \footnotesize
    \begin{tabular}{c c c c}
    \toprule
        \textbf{Loss} & \textbf{Data} & $D_M$ & $D_F$ \\
        \midrule
        \multirow{6}{*}{T} & Adience & \textbf{5.563} & 5.257 \\
        & CelebSET & \textbf{5.532} & 5.478 \\
        & CFD & 8.094 & \textbf{8.238} \\
        & FairFace & \textbf{4.913} & 4.448 \\
        & FARFace & \textbf{8.422} & 4.636 \\
        & UTKFace & \textbf{4.592} & 4.588 \\
        & \blu{LFW} & \blu{4.360} & \blu{\textbf{4.668}} \\
        & \blu{CelebA} & \blu{3.515} & \blu{\textbf{4.039}} \\
        \midrule
        \multirow{6}{*}{A} & Adience & \textbf{10.049} & 9.896 \\
        & CelebSET & \textbf{10.386} & 10.345 \\
        & CFD & 15.375 & \textbf{15.486} \\
        & FairFace & \textbf{8.912} & 8.485 \\
        & FARFace & \textbf{15.482} & 9.269 \\
        & UTKFace & 8.435 & \textbf{8.450} \\
        & \blu{LFW} & \blu{8.003} & \blu{\textbf{8.628}} \\
        & \blu{CelebA} & \blu{7.203} & \blu{\textbf{7.945}} \\
        \midrule
        \multirow{6}{*}{A+T} & Adience & \textbf{5.556} & 5.432 \\
        & CelebSET & \textbf{5.637} & 5.597 \\
        & CFD & 8.186 & \textbf{8.330} \\
        & FairFace & \textbf{4.898} & 4.542 \\
        & FARFace & \textbf{8.455} & 4.721 \\
        & UTKFace & \textbf{4.645} & 4.638 \\
        & \blu{LFW} & \blu{4.358} & \blu{\textbf{4.756}} \\
        & \blu{CelebA} & \blu{3.553} & \blu{\textbf{4.071}} \\
    \bottomrule
    \end{tabular}
    \caption{Average Euclidean distance for LBFC$_{B+R}$.}
    \label{tab:eucl_lbfc_br}
\end{table}

\begin{table}[!t]
    \centering
    \footnotesize
    \begin{tabular}{c c c c}
    \toprule
        \textbf{Loss} & \textbf{Data} & $D_M$ & $D_F$ \\
        \midrule
        \multirow{6}{*}{T} & Adience & \textbf{4.829} & 4.573 \\
        & CelebSET & \textbf{4.948} & 4.943 \\
        & CFD & 7.360 & \textbf{7.470} \\
        & FairFace & \textbf{4.257} & 3.841 \\
        & FARFace & \textbf{7.558} & 4.057 \\
        & UTKFace & 3.988 & \textbf{3.991} \\
        & \blu{LFW} & \blu{3.896} & \blu{\textbf{4.008}} \\
        & \blu{CelebA} & \blu{\textbf{3.220}} & \blu{3.096} \\
        \midrule
        \multirow{6}{*}{A} & Adience & \textbf{6.765} & 6.455 \\
        & CelebSET & 6.889 & \textbf{6.892} \\
        & CFD & 10.253 & \textbf{10.400} \\
        & FairFace & \textbf{6.027} & 5.510 \\
        & FARFace & \textbf{10.496} & 5.820 \\
        & UTKFace & 5.672 & \textbf{5.687} \\
        & \blu{LFW} & \blu{5.514} & \blu{\textbf{5.820}} \\
        & \blu{CelebA} & \blu{4.863} & \blu{\textbf{5.038}} \\
        \midrule
        \multirow{6}{*}{A+T} & Adience & \textbf{5.075} & 4.664 \\
        & CelebSET & \textbf{5.044} & 5.039 \\
        & CFD & 7.515 & \textbf{7.623} \\
        & FairFace & \textbf{4.458} & 3.934 \\
        & FARFace & \textbf{7.814} & 4.145 \\
        & UTKFace & 4.141 & \textbf{4.141} \\
        & \blu{LFW} & \blu{\textbf{4.136}} & \blu{4.098} \\
        & \blu{CelebA} & \blu{\textbf{3.330}} & \blu{3.165} \\
    \bottomrule
    \end{tabular}
    \caption{Average Euclidean distance for LBFC$_{B+R\alpha}$.}
    \label{tab:eucl_lbfc_bra}
\end{table}

\begin{table}[!t]
    \centering
    \footnotesize
    \begin{tabular}{c c c c}
    \toprule
        \textbf{Loss} & \textbf{Data} & $D_M$ & $D_F$ \\
        \midrule
        \multirow{6}{*}{T} & Adience & 4.186 & \textbf{4.296} \\
        & CelebSET & \textbf{4.182} & 4.162 \\
        & CFD & 5.439 & \textbf{5.514} \\
        & FairFace & \textbf{3.767} & 3.630 \\
        & FARFace & \textbf{5.790} & 3.700 \\
        & UTKFace & \textbf{3.775} & 3.771 \\
        & \blu{LFW} & \blu{3.652} & \blu{\textbf{3.857}} \\
        & \blu{CelebA} & \blu{3.137} & \blu{\textbf{3.280}} \\
        \midrule
        \multirow{6}{*}{A} & Adience & 24.221 & \textbf{25.856} \\
        & CelebSET & \textbf{26.519} & 26.190 \\
        & CFD & 36.926 & \textbf{37.570} \\
        & FairFace & 20.708 & \textbf{20.929} \\
        & FARFace & \textbf{35.723} & 22.702 \\
        & UTKFace & \textbf{21.442} & 21.438 \\
        & \blu{LFW} & \blu{19.846} & \blu{\textbf{22.800}} \\
        & \blu{CelebA} & \blu{15.693} & \blu{\textbf{16.190}} \\
        \midrule
        \multirow{6}{*}{A+T} & Adience & 4.205 & \textbf{4.290} \\
        & CelebSET & \textbf{4.219} & 4.195 \\
        & CFD & 5.522 & \textbf{5.581} \\
        & FairFace & \textbf{3.823} & 3.674 \\
        & FARFace & \textbf{5.786} & 3.790 \\
        & UTKFace & \textbf{3.818} & 3.816 \\
        & \blu{LFW} & \blu{3.732} & \blu{\textbf{3.860}} \\
        & \blu{CelebA} & \blu{3.247} & \blu{\textbf{3.391}} \\
    \bottomrule
    \end{tabular}
    \caption{Average Euclidean distance for LBFC$_{B+2}$.}
    \label{tab:eucl_lbfc_b2}
\end{table}

\begin{table}[!t]
    \centering
    \footnotesize
    \begin{tabular}{c c c c}
    \toprule
        \textbf{Loss} & \textbf{Data} & $D_M$ & $D_F$ \\
        \midrule
        \multirow{6}{*}{T} & Adience & \textbf{21.188} & 18.646 \\
        & CelebSET & \textbf{20.688} & 20.279 \\
        & CFD & 29.515 & \textbf{29.814} \\
        & FairFace & \textbf{17.330} & 15.228 \\
        & FARFace & \textbf{33.828} & 15.780 \\
        & UTKFace & \textbf{16.531} & 16.460 \\
        & \blu{LFW} & \blu{15.739} & \blu{\textbf{17.592}} \\
        & \blu{CelebA} & \blu{\textbf{10.902}} & \blu{10.481} \\
        \midrule
        \multirow{6}{*}{A} & Adience & \textbf{24.422} & 22.504 \\
        & CelebSET & \textbf{24.300} & 23.899 \\
        & CFD & 33.741 & \textbf{34.097} \\
        & FairFace & \textbf{20.516} & 18.762 \\
        & FARFace & \textbf{37.753} & 19.571 \\
        & UTKFace & \textbf{19.831} & 19.793 \\
        & \blu{LFW} & \blu{18.819} & \blu{\textbf{20.938}} \\
        & \blu{CelebA} & \blu{\textbf{14.626}} & \blu{14.214} \\
        \midrule
        \multirow{6}{*}{A+T} & Adience & \textbf{21.344} & 19.186 \\
        & CelebSET & \textbf{21.085} & 20.682 \\
        & CFD & 29.853 & \textbf{30.153} \\
        & FairFace & \textbf{17.491} & 15.628 \\
        & FARFace & \textbf{33.992} & 16.119 \\
        & UTKFace & \textbf{16.822} & 16.756 \\
        & \blu{LFW} & \blu{15.886} & \blu{\textbf{18.048}} \\
        & \blu{CelebA} & \blu{\textbf{11.154}} & \blu{10.842} \\
    \bottomrule
    \end{tabular}
    \caption{Average Euclidean distance for LBFC$_{B+2+R}$.}
    \label{tab:eucl_lbfc_b2r}
\end{table}

\begin{table}[!t]
    \centering
    \footnotesize
    \begin{tabular}{c c c c}
    \toprule
        \textbf{Loss} & \textbf{Data} & $D_M$ & $D_F$ \\
        \midrule
        \multirow{6}{*}{T} & Adience & \textbf{7.230} & 6.961 \\
        & CelebSET & \textbf{6.982} & 6.867 \\
        & CFD & 9.736 & \textbf{9.841} \\
        & FairFace & \textbf{6.399} & 5.893 \\
        & FARFace & \textbf{10.492} & 6.056 \\
        & UTKFace & 5.977 & \textbf{5.988} \\
        & \blu{LFW} & \blu{\textbf{6.080}} & \blu{6.002} \\
        & \blu{CelebA} & \blu{4.677} & \blu{\textbf{5.008}} \\
        \midrule
        \multirow{6}{*}{A} & Adience & 10.358 & \textbf{10.478} \\
        & CelebSET & \textbf{10.359} & 10.215 \\
        & CFD & 14.157 & \textbf{14.331} \\
        & FairFace & \textbf{9.321} & 8.886 \\
        & FARFace & \textbf{15.395} & 9.041 \\
        & UTKFace & 8.958 & \textbf{8.982} \\
        & \blu{LFW} & \blu{\textbf{9.082}} & \blu{8.892} \\
        & \blu{CelebA} & \blu{7.487} & \blu{\textbf{7.793}} \\
        \midrule
        \multirow{6}{*}{A+T} & Adience & \textbf{7.039} & 6.924 \\
        & CelebSET & \textbf{6.926} & 6.817 \\
        & CFD & 9.621 & \textbf{9.721} \\
        & FairFace & \textbf{6.222} & 5.841 \\
        & FARFace & \textbf{10.321} & 6.018 \\
        & UTKFace & 5.879 & \textbf{5.889} \\
        & \blu{LFW} & \blu{5.951} & \blu{\textbf{5.986}} \\
        & \blu{CelebA} & \blu{4.572} & \blu{\textbf{4.949}} \\
    \bottomrule
    \end{tabular}
    \caption{Average Euclidean distance for LBFC$_{B+2+R\alpha}$.}
    \label{tab:eucl_lbfc_b2ra}
\end{table}

\begin{table}[!t]
    \centering
    \footnotesize
    \begin{tabular}{c c c c}
    \toprule
        \textbf{Loss} & \textbf{Data} & $D_M$ & $D_F$ \\
        \midrule
        \multirow{6}{*}{T} & Adience & 18.272 & \textbf{18.916} \\
        & CelebSET & \textbf{15.907} & 15.757 \\
        & CFD & \textbf{17.979} & 17.939 \\
        & FairFace & 16.237 & \textbf{17.444} \\
        & FARFace & \textbf{17.535} & 17.324 \\
        & UTKFace & 16.119 & \textbf{16.140} \\
        & \blu{LFW} & \blu{16.626} & \blu{\textbf{16.703}} \\
        & \blu{CelebA} & \blu{\textbf{15.893}} & \blu{15.697} \\
        \midrule
        \multirow{6}{*}{A} & Adience & \textbf{32.986} & 32.878 \\
        & CelebSET & \textbf{19.849} & 19.509 \\
        & CFD & \textbf{25.086} & 24.904 \\
        & FairFace & 31.482 & \textbf{33.344} \\
        & FARFace & \textbf{25.876} & 25.748 \\
        & UTKFace & 32.525 & \textbf{32.543} \\
        & \blu{LFW} & \blu{\textbf{26.301}} & \blu{24.969} \\
        & \blu{CelebA} & \blu{\textbf{21.908}} & \blu{20.727} \\
        \midrule
        \multirow{6}{*}{Cos} & Adience & 27.578 & \textbf{28.406} \\
        & CelebSET & \textbf{20.541} & 20.434 \\
        & CFD & 24.320 & \textbf{24.821} \\
        & FairFace & \textbf{25.632} & 26.828 \\
        & FARFace & \textbf{24.161} & 23.258 \\
        & UTKFace & \textbf{26.103} & 26.051 \\
        & \blu{LFW} & \blu{\textbf{23.704}} & \blu{22.962} \\
        & \blu{CelebA} & \blu{\textbf{22.486}} & \blu{20.575} \\
    \bottomrule
    \end{tabular}
    \caption{Average Euclidean distance for VIT$_{CLS}$.}
    \label{tab:eucl_vit_cls}
\end{table}

\begin{table}[!t]
    \centering
    \footnotesize
    \begin{tabular}{c c c c}
    \toprule
        \textbf{Loss} & \textbf{Data} & $D_M$ & $D_F$ \\
        \midrule
        \multirow{6}{*}{T} & Adience & \textbf{14.772} & 14.722 \\
        & CelebSET & \textbf{15.270} & 15.218 \\
        & CFD & 14.854 & \textbf{14.906} \\
        & FairFace & \textbf{14.552} & 14.488 \\
        & FARFace & \textbf{15.241} & 14.495 \\
        & UTKFace & 14.514 & \textbf{14.524} \\
        & \blu{LFW} & \blu{14.206} & \blu{\textbf{14.466}} \\
        & \blu{CelebA} & \blu{13.223} & \blu{\textbf{13.545}} \\
        \midrule
        \multirow{6}{*}{A} & Adience & 17.158 & \textbf{17.186} \\
        & CelebSET & \textbf{18.993} & 18.745 \\
        & CFD & 17.800 & \textbf{17.936} \\
        & FairFace & \textbf{16.778} & 16.509 \\
        & FARFace & \textbf{20.238} & 15.430 \\
        & UTKFace & 17.242 & \textbf{17.298} \\
        & \blu{LFW} & \blu{16.410} & \blu{\textbf{17.612}} \\
        & \blu{CelebA} & \blu{14.129} & \blu{\textbf{15.229}} \\
        \midrule
        \multirow{6}{*}{Cos} & Adience & \textbf{21.674} & 18.266 \\
        & CelebSET & \textbf{25.381} & 24.799 \\
        & CFD & 22.761 & \textbf{23.237} \\
        & FairFace & \textbf{20.233} & 16.986 \\
        & FARFace & \textbf{28.699} & 17.275 \\
        & UTKFace & 20.099 & \textbf{20.108} \\
        & \blu{LFW} & \blu{\textbf{24.187}} & \blu{18.723} \\
        & \blu{CelebA} & \blu{15.048} & \blu{\textbf{15.845}} \\
        \midrule
        \multirow{6}{*}{A+T} & Adience & \textbf{14.733} & 14.264 \\
        & CelebSET & \textbf{14.990} & 14.880 \\
        & CFD & 14.730 & \textbf{14.786} \\
        & FairFace & \textbf{14.492} & 14.079 \\
        & FARFace & \textbf{15.693} & 14.049 \\
        & UTKFace & 14.476 & \textbf{14.492} \\
        & \blu{LFW} & \blu{\textbf{14.055}} & \blu{13.968} \\
        & \blu{CelebA} & \blu{12.828} & \blu{\textbf{12.935}} \\
        \midrule
        \multirow{6}{*}{Co+A+T} & Adience & \textbf{13.616} & 13.352 \\
        & CelebSET & \textbf{14.344} & 14.281 \\
        & CFD & 14.024 & \textbf{14.081} \\
        & FairFace & \textbf{13.382} & 13.140 \\
        & FARFace & \textbf{14.690} & 13.209 \\
        & UTKFace & 13.457 & \textbf{13.472} \\
        & \blu{LFW} & \blu{13.217} & \blu{\textbf{13.474}} \\
        & \blu{CelebA} & \blu{12.393} & \blu{\textbf{12.738}} \\
    \bottomrule
    \end{tabular}
    \caption{Average Euclidean distance for VIT$_{M}$.}
    \label{tab:eucl_vit_m}
\end{table}

\begin{table}[!t]
    \centering
    \footnotesize
    \begin{tabular}{c c c c}
    \toprule
        \textbf{Loss} & \textbf{Data} & $D_M$ & $D_F$ \\
        \midrule
        \multirow{6}{*}{T} & Adience & \textbf{24.426} & 24.327 \\
        & CelebSET & \textbf{25.198} & 25.060 \\
        & CFD & 25.249 & \textbf{25.314} \\
        & FairFace & 23.557 & \textbf{23.562} \\
        & FARFace & \textbf{25.939} & 24.298 \\
        & UTKFace & 23.370 & \textbf{23.413} \\
        & \blu{LFW} & \blu{23.390} & \blu{\textbf{24.059}} \\
        & \blu{CelebA} & \blu{21.535} & \blu{\textbf{21.716}} \\
        \midrule
        \multirow{6}{*}{A} & Adience & \textbf{27.325} & 26.912 \\
        & CelebSET & \textbf{26.790} & 26.070 \\
        & CFD & 27.478 & \textbf{28.000} \\
        & FairFace & 24.922 & \textbf{25.736} \\
        & FARFace & \textbf{30.054} & 24.276 \\
        & UTKFace & 25.881 & \textbf{25.979} \\
        & \blu{LFW} & \blu{24.670} & \blu{\textbf{26.402}} \\
        & \blu{CelebA} & \blu{20.238} & \blu{\textbf{22.034}} \\
        \midrule
        \multirow{6}{*}{Cos} & Adience & \textbf{34.509} & 33.279 \\
        & CelebSET & \textbf{34.650} & 33.981 \\
        & CFD & 35.711 & \textbf{36.541} \\
        & FairFace & 30.560 & \textbf{30.724} \\
        & FARFace & \textbf{40.118} & 30.873 \\
        & UTKFace & 30.130 & \textbf{30.332} \\
        & \blu{LFW} & \blu{32.158} & \blu{\textbf{32.271}} \\
        & \blu{CelebA} & \blu{27.341} & \blu{\textbf{27.588}} \\
        \midrule
        \multirow{6}{*}{A+T} & Adience & \textbf{21.087} & 20.593 \\
        & CelebSET & \textbf{21.845} & 21.620 \\
        & CFD & 21.618 & \textbf{21.693} \\
        & FairFace & \textbf{20.249} & 19.858 \\
        & FARFace & \textbf{22.792} & 20.493 \\
        & UTKFace & 19.803 & \textbf{19.836} \\
        & \blu{LFW} & \blu{20.777} & \blu{\textbf{20.792}} \\
        & \blu{CelebA} & \blu{\textbf{19.713}} & \blu{19.554} \\
        \midrule
        \multirow{6}{*}{Co+A+T} & Adience & \textbf{22.915} & 22.811 \\
        & CelebSET & \textbf{23.911} & 23.658 \\
        & CFD & 23.558 & \textbf{23.724} \\
        & FairFace & \textbf{22.024} & 21.746 \\
        & FARFace & \textbf{24.840} & 22.293 \\
        & UTKFace & 21.575 & \textbf{21.635} \\
        & \blu{LFW} & \blu{22.474} & \blu{\textbf{22.752}} \\
        & \blu{CelebA} & \blu{\textbf{21.102}} & \blu{20.643} \\
    \bottomrule
    \end{tabular}
    \caption{Average Euclidean distance for VIT$_{CLS+M}$.}
    \label{tab:eucl_vit_cat}
\end{table}

\begin{table}[!t]
    \centering
    \footnotesize
    \begin{tabular}{c c c c}
    \toprule
        \textbf{Loss} & \textbf{Data} & $D_M$ & $D_F$ \\
        \midrule
        \multirow{6}{*}{T} & Adience & 10.077 & \textbf{12.524} \\
        & CelebSET & 12.243 & \textbf{12.308} \\
        & CFD & 12.595 & \textbf{12.737} \\
        & FairFace & \textbf{10.968} & 10.593 \\
        & FARFace & \textbf{13.529} & 10.987 \\
        & UTKFace & \textbf{11.550} & 11.526 \\
        & \blu{LFW} & \blu{11.256} & \blu{\textbf{12.106}} \\
        & \blu{CelebA} & \blu{10.462} & \blu{\textbf{10.830}} \\
        \midrule
        \multirow{6}{*}{A} & Adience & 14.040 & \textbf{16.225} \\
        & CelebSET & 15.771 & \textbf{15.834} \\
        & CFD & 16.490 & \textbf{16.631} \\
        & FairFace & \textbf{15.622} & 14.735 \\
        & FARFace & \textbf{17.888} & 15.010 \\
        & UTKFace & \textbf{15.536} & 15.496 \\
        & \blu{LFW} & \blu{14.609} & \blu{\textbf{14.766}} \\
        & \blu{CelebA} & \blu{13.268} & \blu{\textbf{13.451}} \\
        \midrule
        \multirow{6}{*}{Cos} & Adience & 75.745 & \textbf{78.757} \\
        & CelebSET & 82.551 & \textbf{82.964} \\
        & CFD & 82.158 & \textbf{83.226} \\
        & FairFace & \textbf{81.578} & 64.050 \\
        & FARFace & \textbf{110.671} & 58.004 \\
        & UTKFace & \textbf{78.920} & 78.233 \\
        & \blu{LFW} & \blu{\textbf{80.738}} & \blu{70.964} \\
        & \blu{CelebA} & \blu{\textbf{67.583}} & \blu{64.406} \\
        \midrule
        \multirow{6}{*}{A+T} & Adience & 9.938 & \textbf{12.611} \\
        & CelebSET & 12.215 & \textbf{12.287} \\
        & CFD & 12.625 & \textbf{12.771} \\
        & FairFace & \textbf{10.859} & 10.659 \\
        & FARFace & \textbf{13.501} & 11.019 \\
        & UTKFace & \textbf{11.546} & 11.525 \\
        & \blu{LFW} & \blu{11.149} & \blu{\textbf{12.052}} \\
        & \blu{CelebA} & \blu{10.271} & \blu{\textbf{10.717}} \\
        \midrule
        \multirow{6}{*}{Co+A+T} & Adience & 10.298 & \textbf{12.916} \\
        & CelebSET & 12.551 & \textbf{12.635} \\
        & CFD & 12.988 & \textbf{13.127} \\
        & FairFace & \textbf{11.303} & 10.984 \\
        & FARFace & \textbf{13.943} & 11.433 \\
        & UTKFace & \textbf{11.874} & 11.853 \\
        & \blu{LFW} & \blu{11.416} & \blu{\textbf{12.300}} \\
        & \blu{CelebA} & \blu{10.549} & \blu{\textbf{10.991}} \\
    \bottomrule
    \end{tabular}
    \caption{Average Euclidean distance for IBLIP.}
    \label{tab:eucl_iblip}
\end{table}

\end{document}